%% file: main.tex
\documentclass[11pt]{article}

\usepackage[]{acl}

\usepackage{times}
\usepackage{latexsym}

\usepackage[T1]{fontenc}

\usepackage[utf8]{inputenc}

\usepackage{microtype}

\usepackage{inconsolata}

\usepackage{graphicx}

\usepackage{amsmath}
\usepackage{multirow}

\usepackage{tcolorbox}

\usepackage{booktabs}
\usepackage{tabularx}
\usepackage[table]{xcolor}
\usepackage{rotating}
\usepackage{enumitem}
\usepackage{multirow}
\usepackage{array}
\tcbuselibrary{breakable}
\usepackage{pdflscape}
\usepackage{makecell}
\usepackage{graphicx}
\usepackage{caption}
\usepackage{multicol}

\usepackage{graphicx}
\usepackage{subcaption}
\usepackage{booktabs}
\usepackage[section]{placeins}
\usepackage[font=small,labelfont=bf]{caption}
\newtcolorbox{prompttemplate}{
  colback=white,
  colframe=black,
  boxrule=0.8pt,
  arc=2mm,
  left=2mm,right=2mm,top=1.5mm,bottom=1.5mm,
  breakable
}

\title{How Do Prompt Variations Affect Energy Consumption in On-Device LLMs?}

\author{
  \textbf{Wei Hu\textsuperscript{1}\thanks{\ Equal contribution.}},
  \textbf{Xiaolong Tu\textsuperscript{1}\footnotemark[1]},
  \textbf{Dawei Chen\textsuperscript{2}},
  \textbf{Yitao Chen\textsuperscript{2}},
  \textbf{Kyungtae Han\textsuperscript{2}},
  \textbf{Haoxin Wang\textsuperscript{1}\thanks{\ Corresponding author.}}
\\
  \textsuperscript{1}Georgia State University \quad
  \textsuperscript{2}Toyota Motor North America
\\
  \small{
    \texttt{\{whu6, xtu1\}@student.gsu.edu, haoxinwang@gsu.edu}
  }
\\
  \small{
    \texttt{\{dawei.chen1, yitao.chen, kt.han\}@toyota.com}
  }
}

\begin{document}

\maketitle
\begin{abstract}
Large language models (LLMs) are increasingly deployed on mobile devices, making energy efficiency a key deployment constraint, yet the energy impact of prompt design remains underexplored. This paper aims to understand how two prompt properties, cognitive load and phrasing pattern, shape the energy behavior of on-device LLM inference. We conduct a broad empirical study covering prompt properties, datasets, models, and devices, with phase-level profiling that separates prefill and decode energy. We find that cognitive load primarily affects the energy cost per token, while phrasing pattern affects energy largely through token usage. Our energy-quality analysis further shows that prompt design reshapes the attainable frontier differently across models, highlighting the need for model-aware prompt design in energy-efficient on-device LLM inference. Code, datasets, and scripts are available at \url{https://amai-gsu.github.io/PromptProperty/}.
\end{abstract}

\section{Introduction}
The deployment of Large Language Models (LLMs) is increasingly expanding beyond cloud serving toward on-device execution on mobile and edge platforms \cite{10.1145/3398209, 8736011}. This trend is driven by the growing demand for stronger user privacy guarantees, lower inference latency, and reliable offline access \cite{10.1145/3712001}. However, unlike cloud-based inference, where energy consumption is amortized across large-scale infrastructure and has little immediate impact on users, on-device inference is powered by finite batteries, making energy efficiency a strict and user-visible constraint \cite{3692070.3693386}. As a result, energy efficiency has emerged as a key bottleneck limiting widespread adoption of on-device LLMs \cite{10.1145/3381831}.

Prior research on efficient LLMs has overwhelmingly focused on model-centric optimization techniques, including quantization, pruning, architectural compression, and distillation \cite{pmlr-v202-xiao23c,NEURIPS2022_c3ba4962,frankle2018the,ICLR2024_8ac015d4,hinton2015distilling}. 
These approaches improve inference efficiency primarily by reducing model size and computational cost through parameter or architectural modification, and have become the dominant strategy for on-device LLM deployment. 
In addition, several recent studies have examined prompt sensitivity by analyzing how variations in prompt phrasing, structure, and formatting affect the quality and stability of model outputs \cite{long-etal-2025-makes,ismithdeen-etal-2025-promptception,10.1145/3689217.3690621, lu-etal-2022-fantastically}.
However, a fundamental question remains largely unexplored: \textit{how does prompt design itself affect the energy consumption of on-device LLM inference?} If different prompt variations induce substantially different energy costs, even when producing comparable outputs, prompting could serve as a new, model-agnostic approach to improving energy efficiency beyond model-centric optimization.

Studying how prompt variations affect the energy consumption of on-device LLMs is non-trivial. First, prompts are linguistic artifacts, whereas energy consumption depends on the computation they trigger inside the model, and there is no direct mapping between the two. Understanding why a prompt is energy-consuming therefore requires carefully designed empirical analysis that systematically links prompt variations to differences in computational behavior and measured power consumption. Second, it is challenging to construct prompt variants in a principled way. Prompt variants can differ along many dimensions, such as phrasing, cognitive load, or length, and na\"{\i}ve changes may unintentionally alter task semantics. Designing prompt variants that isolate specific prompt properties while preserving comparable task intent is essential for attributing observed energy differences to prompt design.
Finally, deriving generalizable insights requires moving beyond isolated examples to identify consistent patterns that hold across prompt variants, tasks, models, and hardware devices.

To address these challenges, we design and conduct a comprehensive empirical study that systematically examines the impact of prompt variants on the energy consumption of on-device LLM inference. Our contributions are summarized as follows:
\begin{itemize}
    
    \item \textbf{New dataset construction for cognitive load.} We construct a new prompt dataset that varies cognitive demand while preserving task intent. We then develop LLM-based scoring and embedding similarity to ensure semantic consistency and limit semantic variation. This dataset enables a controlled examination of energy variation associated with cognitive load demand rather than semantic differences.

    \item \textbf{Empirical study of prompt property effects.} We present the first large-scale empirical study of how two prompt properties, \textit{Phrasing Pattern} and \textit{Cognitive Load}, affect energy consumption in on-device LLM inference, spanning an evaluation space of prompt variants $\times$ datasets $\times$ models $\times$ devices. This enables \textit{the first study of how linguistic form and reasoning demand influence the computational behavior of on-device LLM inference.}
    
    \item \textbf{Energy behavior analysis and empirical findings.} We analyze energy behavior using per-token energy and token usage, which indicate the cost per token and the number of tokens processed or generated, respectively. We find that cognitive load mainly affects per-token energy variation, while phrasing patterns primarily affect token usage. We further show that energy-quality trade-offs are model-dependent, offering guidance for model-aware, energy-efficient prompt design.
\end{itemize}

\section{Prompt Properties and Datasets}
\label{sec:property_dataset}

 This section first describes the two prompt properties motivated by prior literature, and then introduces the datasets used to study them. 

\subsection{Prompt Properties}
\label{subsec:method_promptsets}

\begin{figure*}[t]
  \centering
  \includegraphics[width=\linewidth]{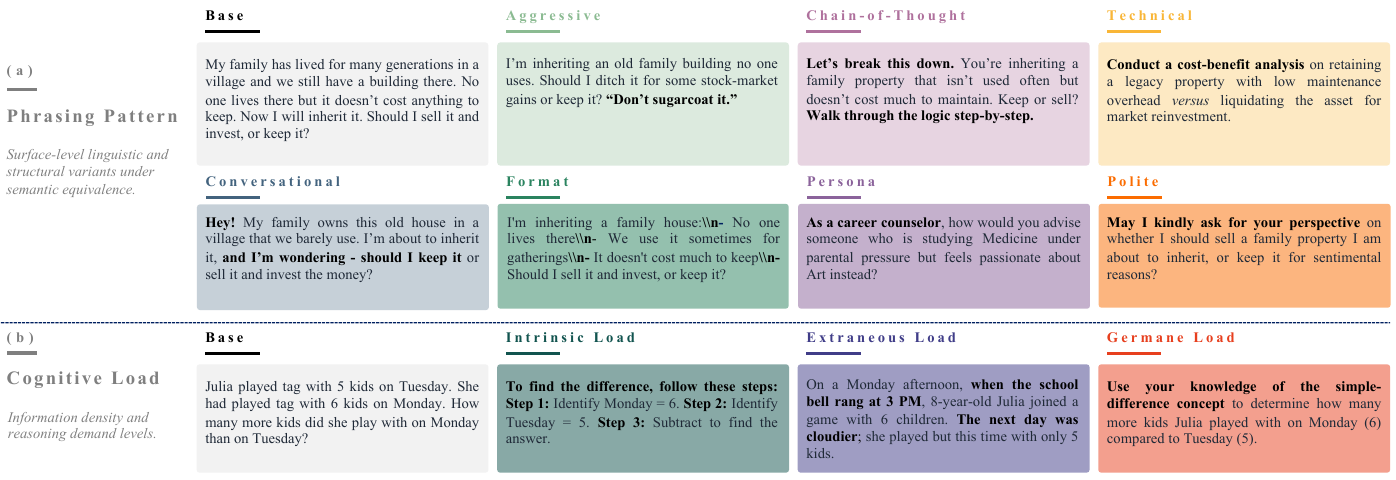}
  \caption{\textbf{Prompt examples for two prompt properties.} (a) Phrasing pattern includes a base prompt and seven sub-properties that capture surface-level linguistic and structural variations under the same semantic intent. (b) Cognitive load includes a base prompt and three sub-properties that vary the information density and reasoning demand of the prompt.}
  \label{fig:prompt_property_examples}
\end{figure*}

To analyze how prompt variations affect energy behavior, we focus on two representative prompt properties: \textit{Phrasing Pattern} and
\textit{Cognitive Load}. Phrasing pattern characterizes linguistic and structural variation under semantic equivalence, while cognitive load introduces the reasoning structure and cognitive demand. The two properties capture complementary aspects of prompt variation, covering both expression form and reasoning demand, as shown in Figure~\ref{fig:prompt_property_examples}. The detailed definitions of the phrasing pattern and cognitive load sub-properties are provided in Appendix~\ref{app:prompt_property_definitions}.

\noindent \textbf{Phrasing pattern.}
Prior studies have shown that semantically equivalent paraphrases can induce substantial variability in model behavior, raising concerns about robustness and evaluation reliability~\cite{ismithdeen-etal-2025-promptception,10.1162/tacl_a_00681}. Motivated by this sensitivity, we study phrasing pattern as a prompt property that characterizes variations in tone, linguistic style, and presentation structure while maintaining semantic intent. Drawing on the taxonomy of phrasing styles introduced by \citet{sotic2025university} for the CLEF 2025 ELOQUENT Lab \cite{karlgren2025overview}, we consider seven sub-properties:
\textit{Aggressive Tone}, \textit{Conversational Tone},
\textit{Chain-of-Thought (CoT)}, \textit{Formatting Differences},
\textit{Persona-Based Prompts}, \textit{Polite Tone}, and
\textit{Technical/Jargon-Heavy Prompts}.

\noindent \textbf{Cognitive load.}
Cognitive Load Theory categorizes cognitive load into intrinsic load, extraneous load, and
germane load, emphasizing that problem-solving requires meticulous management of cognitive load \cite{Sweller01121991}. Recent work conceptualizes cognitive load as a prompt-level property that systematically influences model reasoning behavior \cite{long-etal-2025-makes}. 
In this paper, we examine cognitive load through three sub-properties: \textit{Intrinsic Load}, \textit{Extraneous Load},
and \textit{Germane Load}. Each sub-property is represented by explicit prompt cues that vary the reasoning demand while preserving the original task intent. 

\subsection{Prompt Datasets}
\label{subsec:method_datasets}

For phrasing pattern, we select variants from an existing robustness evaluation dataset; for cognitive load, we construct a new dataset.

\noindent \textbf{Public benchmark dataset for phrasing pattern.}
To study phrasing pattern, we use the dataset introduced by \citet{sotic2025university}, which offers controlled stylistic variations of semantically equivalent prompts for robustness evaluation. We retain prompts covering the seven selected sub-properties.

\noindent \textbf{New dataset construction for cognitive load.}
As no existing public benchmark is designed to capture cognitive load across its three sub-properties (intrinsic, extraneous, and germane load), we construct a new dataset for cognitive load property using a manually designed template that guides the LLM in generating prompt variants (see Appendix~\ref{app:cogload_template}). For each sub-property, we create dedicated prompt variants while minimizing confounding cues from the other sub-properties. We sample base prompts as semantic anchors from three public datasets that represent distinct reasoning types: SVAMP (arithmetic word problems) \cite{patel-etal-2021-nlp}, BoolQ (binary yes/no question answering) \cite{clark-etal-2019-boolq}, and AI2-ARC (multiple-choice science questions) \cite{clark2018thinksolvedquestionanswering}.

\section{Empirical Study Pipeline}
\label{sec:implementation}

\begin{figure*}[t]
  \centering
  \includegraphics[width=\linewidth]{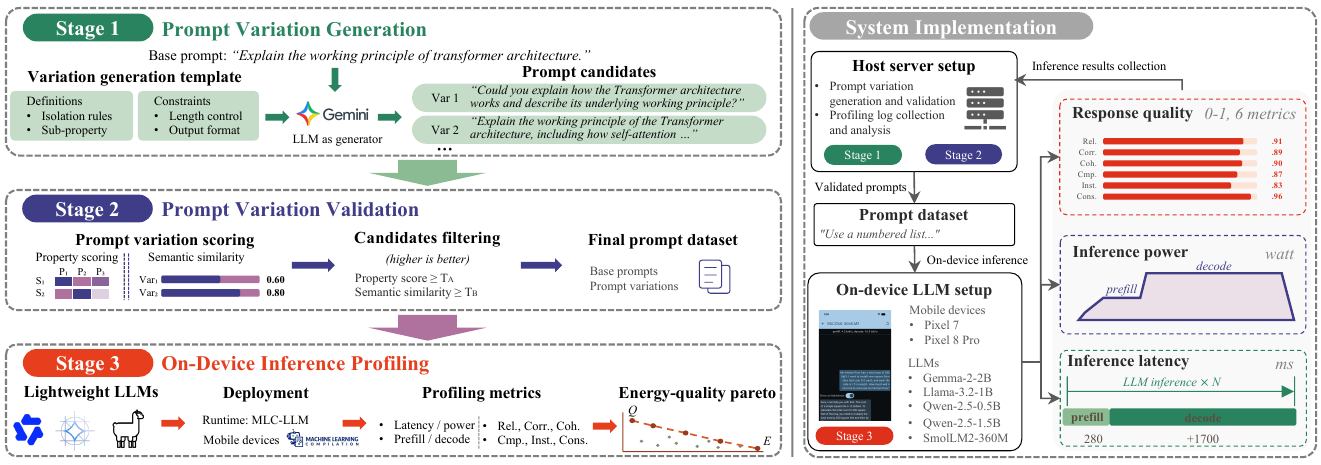}
  \caption{
\textbf{Empirical study pipeline.} Prompt variants are generated from base prompts using property definitions and generation constraints, then validated through scoring and semantic similarity filtering. Validated prompts are deployed on lightweight LLMs (Gemma-2-2B \cite{team2024gemma}, Llama-3.2-1B \cite{grattafiori2024llama}, Qwen-2.5-0.5B/1.5B \cite{qwen2.5},  and SmolLM2-360M \cite{allal2025smollm2}) via MLC-LLM to collect phase-level power, latency, and response quality measurements.}
  \label{fig:impl_pipeline}
\end{figure*}

This section describes the empirical study pipeline we designed. As shown in Figure~\ref{fig:impl_pipeline}, the workflow consists of three stages: prompt generation, prompt validation, and on-device energy profiling.

\subsection{Prompt Variation Generation}
\label{sec:prompt_generator}

For cognitive load property, we generate three variants for each base prompt, corresponding to intrinsic, extraneous, and germane load. We employ Gemini-2.5-Pro API \cite{comanici2025gemini} as the generator. Given a base prompt sampled from public datasets, the generator produces candidate prompt variations that target a specific prompt sub-property while preserving semantic meaning. The generation process follows three rules: (i) \textit{Single-property isolation}, where each variation instantiates only one prompt sub-property to avoid property overlap;
(ii) \textit{Semantic preservation}, where the original task intent and expected answer remain unchanged; and
(iii) \textit{Sub-property consistency}, where each individual variation strictly follows the precise definition of its intended target sub-property.

\subsection{Prompt Variation Validation}
\label{sec:prompt_validation}

To ensure that the generated prompt variants are valid and aligned with the intended cognitive load sub-property, all candidates are passed through a rigorous validation process: (i) \textit{Property purity}: confirming that each candidate strictly reflects its target sub-property while avoiding overlap with other sub-properties. (ii) \textit{Semantic similarity}: verifying that each candidate remains semantically aligned with the base prompt, with alignment quantified using embedding-based cosine similarity.

\noindent \textbf{Prompt variation scoring.}
For prompt variation quality control, we adopt the rubric-based scoring system from \cite{long-etal-2025-makes}, which provides not only the definitions but also the standardized evaluation template for cognitive load properties (see Appendix~\ref{app:prompt_evaluator}). Specifically, we invert the standard scoring direction for extraneous load, so that a higher score indicates a stronger presence of irrelevant and redundant information and better alignment with the intended extraneous load sub-property.

Following the evaluation template, each variation receives a score vector
$\mathbf{s} = (s_i, s_e, s_g)$ over the three cognitive load sub-properties. To isolate the
target property, we retain only variations whose target sub-property score is at least
8 while all non-target scores are at most 3 on the 1-10 scale.

\noindent \textbf{Semantic consistency filtering.}
We further filter each prompt variation group to prevent
semantic drift from the original task. Specifically, for each base prompt embedding
$\mathbf{e}_0$ and variation embedding $\mathbf{e}_i$, we compute cosine similarity.
Given the structural rewriting required by cognitive load instantiation, we adopt
a threshold $\tau$ and retain a prompt variation group only if all its
variations remain above this threshold:
\begin{equation}
    \min_i \mathrm{CosSim}(\mathbf{e}_0,\mathbf{e}_i) \ge \tau,
\end{equation}
where we use a relaxed threshold $\tau=0.6$ to accommodate added cognitive load information.

\subsection{On-Device LLM Inference Profiling}
\label{sec:profiling}

We run the validated prompts on mobile devices using instruction-tuned,
quantized models deployed locally through the MLC framework. All prompts are
evaluated using a fixed inference configuration.

\noindent \textbf{On-device power and latency profiling framework.}
Building on MLC-LLM \cite{mlcllm2023} and LM-Meter \cite{10.1145/3769102.3770614}, we develop a profiling framework for measuring power and latency during on-device LLM inference. The framework instruments the inference runtime to record the start ($t_s$) and end ($t_e$) timestamps of the prefill and decode phases, which define phase-level latency and the corresponding energy integration  windows. We divide on-device inference into two phases. The prefill window spans from request arrival to first-token sampling, covering tokenization, queueing, embedding, the prefill forward pass, and first-token sampling; the decode window covers all subsequent token generation. This extended window is used instead of the isolated prefill forward pass to capture all energy consumed before the first output and cleanly separate the two phases.

Concurrently, we sample device-level current $I(t)$ and voltage $V(t)$ through Android Debug Bridge (ADB) to construct power traces. Hardware-specific interfaces and device settings are summarized in Table~\ref{tab:exp_setup}. Instantaneous power is computed as $P(t)=I(t)\cdot V(t)$. After aligning the inference traces with the power measurements, we estimate the phase-level energy $E_{\text{phase}}$ by integrating power over the corresponding phase window $[t_s,t_e]$, approximated using discrete samples:

\begin{equation}
E_{\text{phase}}=\int_{t_s}^{t_e} P(t)\,dt \approx \sum_{i\in\mathcal{I}_{\text{phase}}} P_i \Delta t ,
\end{equation}

\noindent
where $\mathcal{I}_{\text{phase}}$ denotes the set of power-sample indices whose timestamps fall within the phase window, and $\Delta t$ is the effective sampling interval.

\noindent \textbf{Response logging.}
For each run, we record the generated response and run-level metadata, including device, model, prompt ID, token counts, and timestamps. These records support traceable analysis across prompts, models, devices, and energy measurements. Detailed latency, energy, token, and quality statistics are provided in Appendix~\ref{app:experimental_results}.

\subsection{Response Quality Evaluation}
\label{subsec:eval_response_quality}
We evaluate response quality with task-specific metrics: accuracy for ground-truth tasks and reference-free assessment for open-ended tasks using DeepEval \cite{Ip_deepeval_2026}, with a randomly sampled subset manually verified.

\noindent \textbf{Cognitive load tasks.}
For cognitive load datasets with ground-truth answers, we measure response quality with Exact Match (EM) accuracy. Since prompt variations preserve semantic intent, models should produce the same correct answer across variants. EM helps ensure that energy differences are not attributable to degraded correctness.

\noindent \textbf{Phrasing pattern tasks.}
For phrasing pattern tasks, strict accuracy is unsuitable because the prompts are open-ended and can elicit multiple valid responses. In addition, the reference answers are LLM-generated rather than objective ground truth. We score each open-ended response along six dimensions:
\textit{Relevance}, \textit{Correctness}, \textit{Coherence}, \textit{Completeness},
\textit{Instruction Adherence}, and \textit{Internal Consistency}. Each dimension is scored on a 0-1 rubric, with the dimension definitions and evaluation rubric provided in Appendix~\ref{app:metrics}.

\section{Results and Analysis}
\label{sec:results_analysis}

Prefill processes the input prompt, while decode generates the response. We therefore report phase-level energy alongside total energy to analyze how prompt variations affect each phase. We compare per-token energy with token usage, indicating the cost of processing each token and the number of tokens processed or generated:
\begin{equation}
E_{\text{phase}} = e_{\text{phase}} \cdot T_{\text{phase}},
\label{eq:energy_factorization}
\end{equation}
where $E_{\text{phase}}$ denotes the energy consumption of a given inference phase, $e_{\text{phase}}$ denotes the per-token energy, and $T_{\text{phase}}$ denotes the number of tokens processed or generated in that phase.

To understand how prompt properties affect the \emph{energy cost per token} and the \emph{number of tokens processed and generated}, we organize this section around the following three research questions:

\begin{itemize}
    \item \textbf{\textit{RQ1: To what extent is per-token energy shaped by prompt properties relative to device and model factors?}}
    \item \textbf{\textit{RQ2: How do prompt properties affect token usage across the prefill and decode phases?}}
    \item \textbf{\textit{RQ3: Can prompt properties shift the energy-quality frontier across model architectures?}}
\end{itemize}

\begin{figure*}[t]
    \centering
    \begin{subfigure}[t]{0.49\linewidth}
        \centering
        \includegraphics[width=\linewidth]{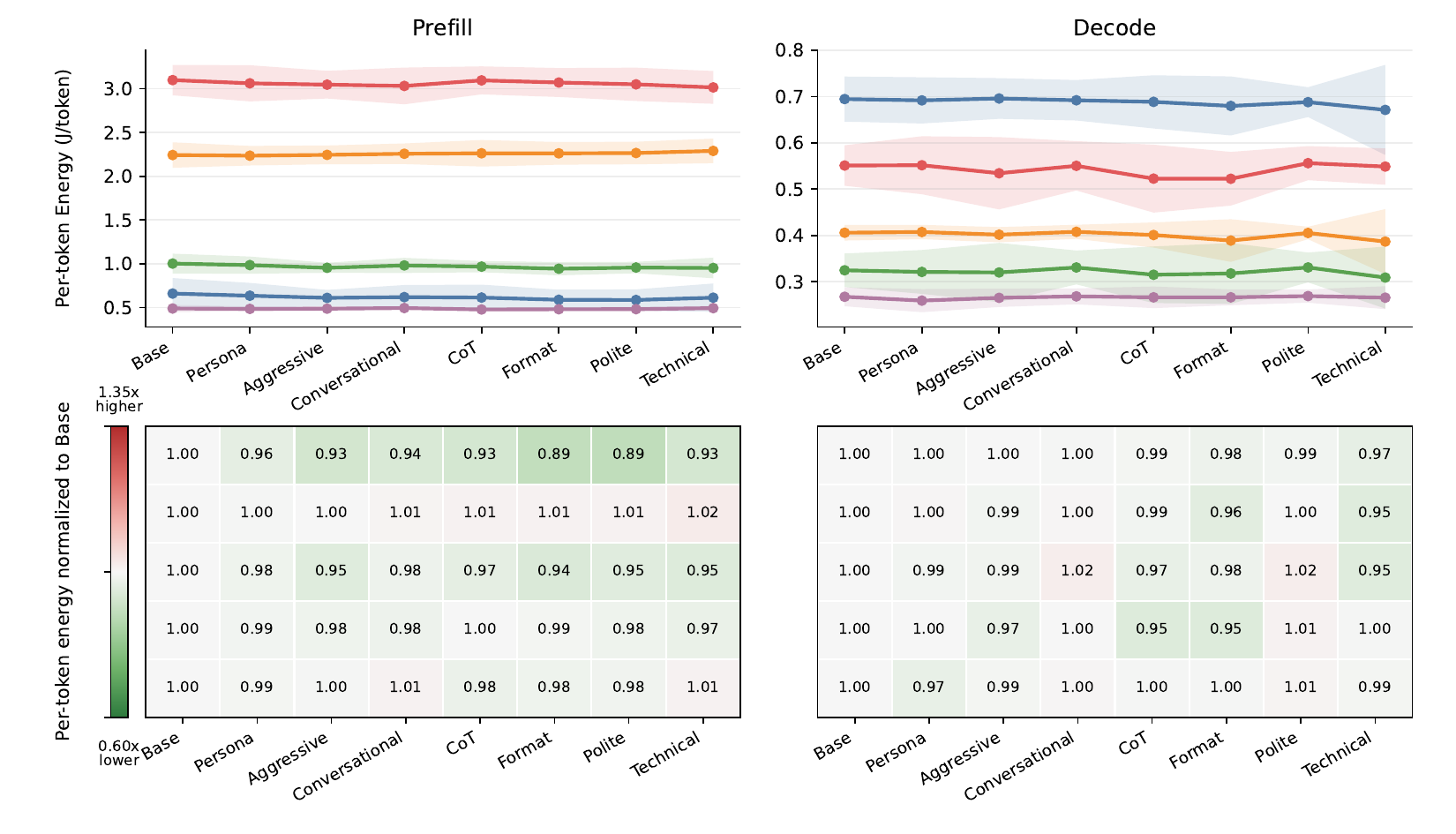}
        \caption{Per-token energy across phrasing pattern for Pixel 8 Pro.}
        \label{fig:rq1_per_token_energy_phrasing}
    \end{subfigure}
    \hfill
    \begin{subfigure}[t]{0.49\linewidth}
        \centering
        \includegraphics[width=\linewidth]{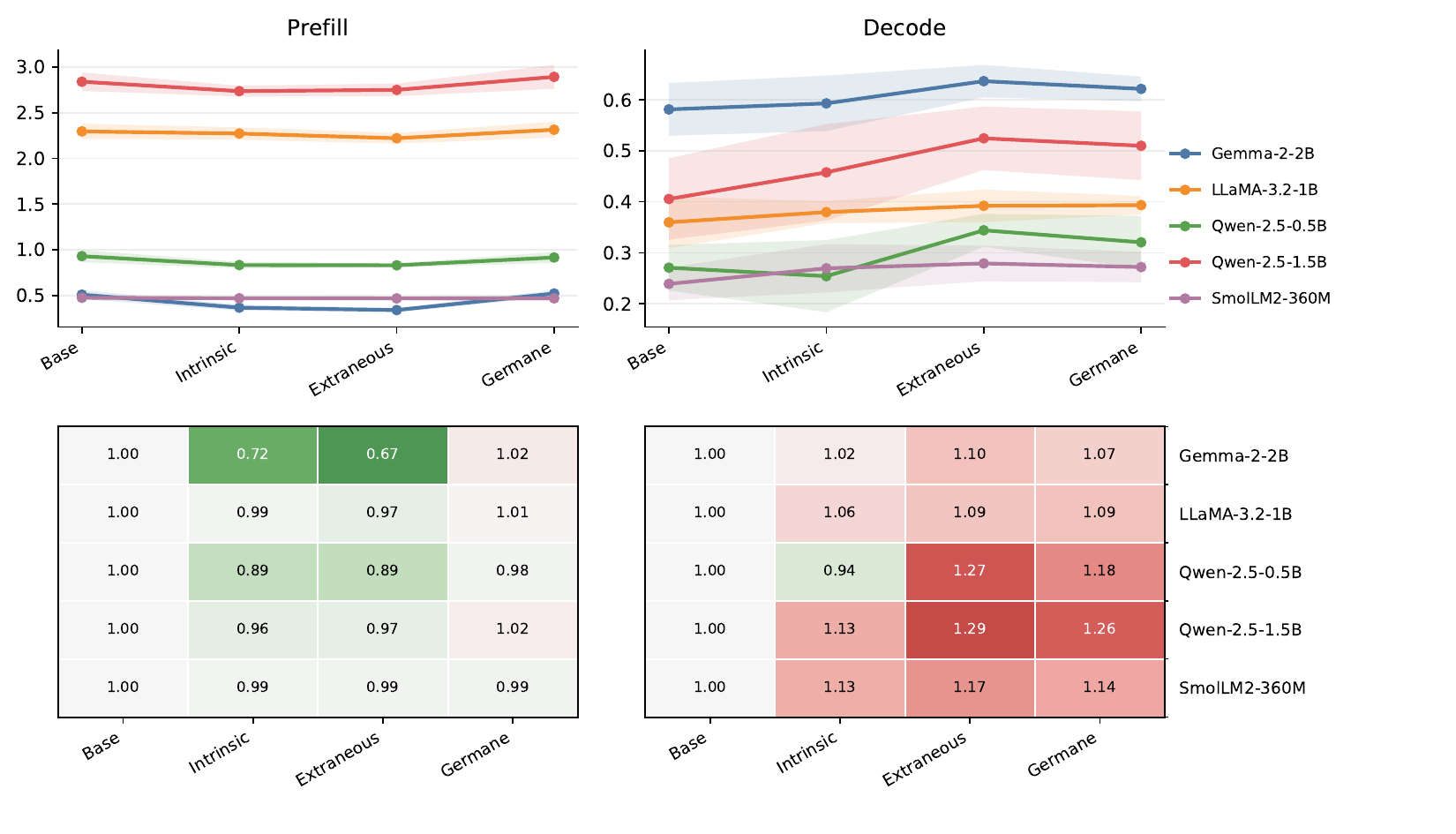}
        \caption{Per-token energy across cognitive load for BoolQ.}
        \label{fig:rq1_per_token_energy_cognitive}
    \end{subfigure}

    \caption{\textbf{Per-token energy analysis across prompt properties.} (a) Phrasing pattern results on Pixel 8 Pro across prefill and decode phases, and (b) cognitive load results on BoolQ across prefill and decode phases. The first row shows absolute per-token energy for the prefill and decode phases, with shaded bands denoting $\pm 1$ standard deviation across prompts within each sub-property; the second row reports each sub-property's per-token energy normalized by the corresponding base prompt.}
    \label{fig:rq1_per_token_energy}
\end{figure*}

\subsection{RQ1: Per-token Energy Analysis}
\label{sec:rq1_per_token_energy}

To answer RQ1, we first examine whether prompt properties change the energy cost of processing each token, as shown in Figure~\ref{fig:rq1_per_token_energy}. Due to space constraints, we provide the remaining per-token energy analyses across prompt properties, devices, and models in Appendix~\ref{app:rq1_additional_results}.

\noindent\textit{Observation 1: Prefill costs more energy per token than decode on
most models.} On LLaMA-3.2-1B and both Qwen models, per-token prefill
energy is 3-6$\times$ that of decode, while Gemma-2-2B shows the
opposite. This inversion relative to server-scale inference is known for
on-device runtimes, reflecting limited compute, memory bandwidth, and the
absence of parallelized prefill support \cite{10.1145/3769102.3770614}.
Part of the prefill window (e.g., tokenization, queueing, embedding) does
not scale with prompt length; thus on-device prefill exhibits a pronounced
amortization effect: longer prompts yield lower per-token energy. The
effect is largest on Gemma-2-2B, which has the lowest marginal per-token
prefill energy: its normalized prefill values fall to 0.67× under
extraneous load.

\noindent\textit{Observation 2: Absolute per-token energy is primarily model-dependent.}
Across both prompt properties, absolute per-token energy is primarily determined by the executed model rather than the prompt sub-property. Across both phrasing pattern and cognitive load, the separation between model curves is much larger than the variation across prompt sub-properties within the same model. This indicates that model architecture and scale remain the dominant factors in per-token execution cost.

\noindent\textit{Observation 3: Phrasing pattern exhibits limited normalized per-token energy variation.}
For phrasing pattern variants, the normalized heatmaps remain close to the base prompt across most models and sub-properties. Most values fluctuate around 1.0 in both prefill and decode, indicating that surface-level changes in tone, persona, formatting, politeness, or technical wording do not substantially alter the cost of processing each token. While some models show small deviations for specific sub-properties, these changes are modest compared with the absolute differences across models.

\noindent\textit{Observation 4: Cognitive load substantially affects per-token decode energy.}
Compared with phrasing pattern, cognitive load variants produce more visible changes in per-token energy in decode phases. During decoding, extraneous and germane load increase per-token energy for Qwen-2.5-0.5B, Qwen-2.5-1.5B, and SmolLM2-360M, with extraneous load reaching 1.29$\times$ on Qwen-2.5-1.5B. This indicates that cognitive load affects per-token execution cost in a phase- and model-dependent manner.

\begin{tcolorbox}[
    colback=green!5!white,
    colframe=teal!70!black,
    boxrule=0.7pt,
    arc=3mm,
    left=1mm,
    right=1mm,
    top=1mm,
    bottom=1mm
]
\textit{\textbf{Finding.}} Per-token energy is primarily model-driven, but cognitive load introduces variation in per-token inference cost that surface phrasing largely does not. This suggests that energy analysis should distinguish reasoning demand from linguistic form, rather than relying only on aggregate energy or token-count explanations.
\end{tcolorbox}

\begin{figure*}[t]
    \centering
    \includegraphics[width=\linewidth]{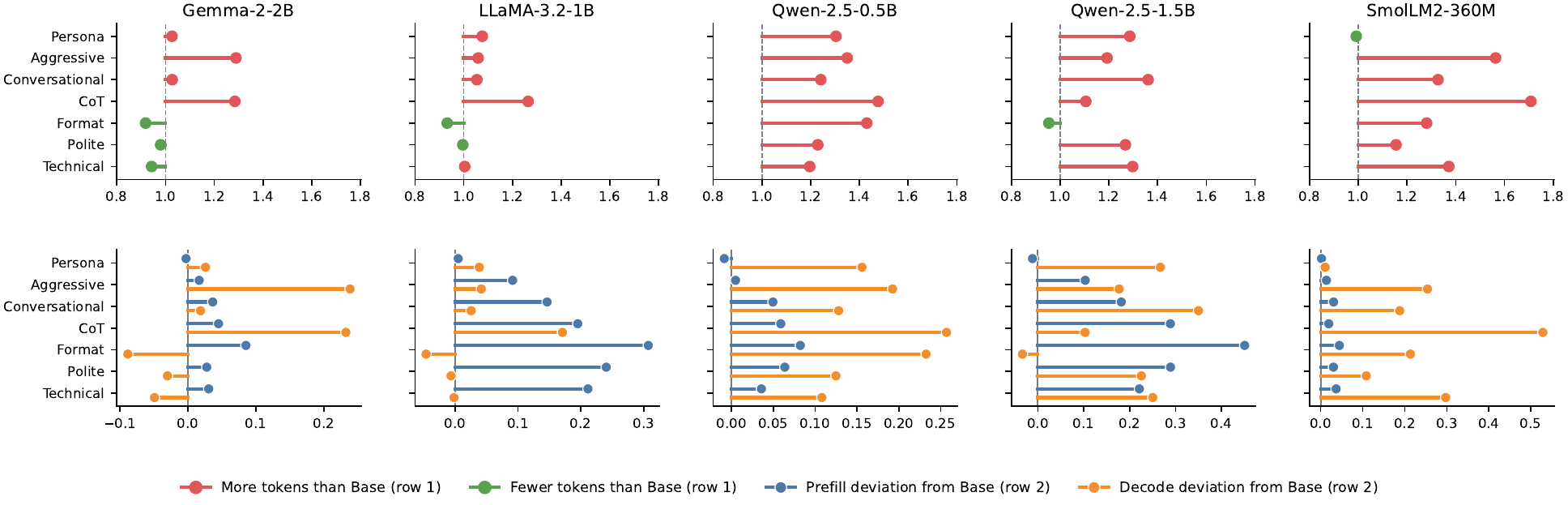}
    \caption{
\textbf{Token length and fixed-baseline phase-level energy burden for phrasing pattern on Pixel 8 Pro.}
Top: total token ratio relative to Base, computed as each sub-property token count divided by that of the corresponding Base prompt (>1: more tokens; <1: fewer tokens). Bottom: fixed-baseline prefill/decode burden relative to Base, normalized by base prompt/completion tokens, respectively and after subtracting the corresponding Base burden; positive/negative values indicate higher/lower burden.}
\label{fig:rq2_combined_tokenratio_phaseburden}
\end{figure*}

\subsection{RQ2: Token Footprint Analysis}
\label{sec:rq2_token_usage}

As shown in Figure~\ref{fig:rq2_combined_tokenratio_phaseburden}, we analyze how phrasing patterns affect token usage and phase-level burden. Due to space constraints, we provide the remaining token usage and phase burden results across properties, devices, and models in Appendix~\ref{app:rq2_additional_results}.

\noindent\textit{Observation 1: Phrasing patterns substantially reshape token usage.} The top row shows that CoT and aggressive prompts consistently increase token usage relative to Base, while format, polite, and technical prompts stay closer to Base and sometimes reduce tokens. Therefore, surface-level phrasing alone can alter inference computation, even when per-token energy shifts are small.

\noindent\textit{Observation 2: Token usage effects are model dependent.}
The same phrasing pattern can lead to different token expansion across models. For example, CoT produces much larger total token ratios on SmolLM2-360M and Qwen-2.5-0.5B than on Gemma-2-2B or LLaMA-3.2-1B. This suggests that phrasing patterns affect not only the input prompt but also model generation behavior.

\noindent\textit{Observation 3: Phase burden differs across phrasing patterns.}
The bottom row shows that token-related energy burden is not distributed uniformly across inference phases. CoT and aggressive prompts often increase decode burden, indicating stronger effects on response generation. In contrast, format prompts can increase prefill burden while reducing or weakly increasing decode burden, suggesting that some phrasing patterns shift cost toward input processing rather than generation.

\begin{tcolorbox}[
    colback=green!3!white,
    colframe=teal!80!black,
    boxrule=0.7pt,
    arc=3mm,
    left=4mm,
    right=4mm,
    top=2mm,
    bottom=2mm
]
\textit{\textbf{Finding.}} Phrasing patterns shape energy behavior primarily by affecting token usage during inference, rather than the per-token processing cost. This suggests that energy consumption does not simply scale with total token usage; energy analysis should also consider how phrasing patterns alter model-specific generation behavior.
\end{tcolorbox}

\begin{figure*}[t]
    \centering
    \includegraphics[width=\linewidth]{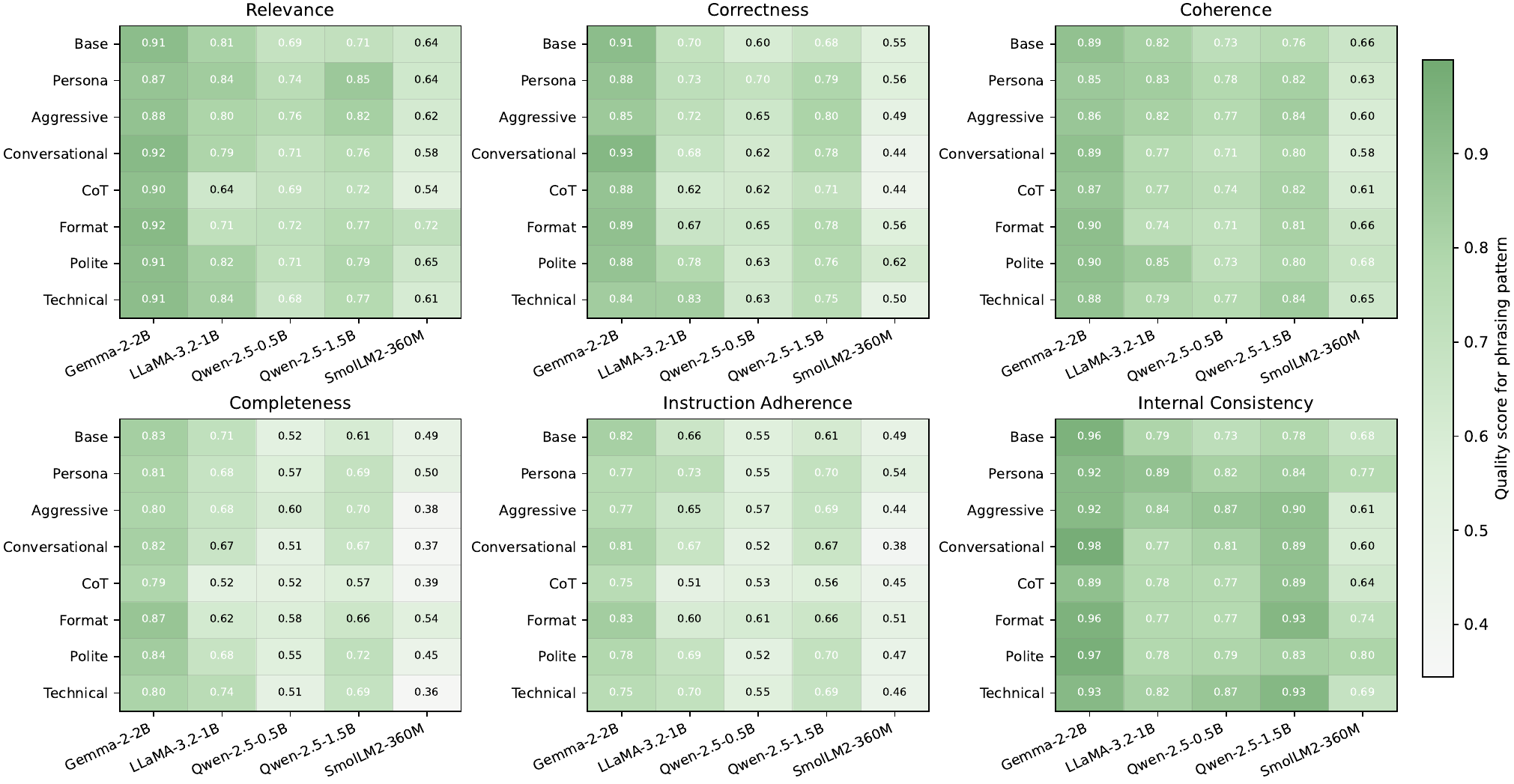}
    \caption{\textbf{Response quality across phrasing pattern sub-properties and models.} Heatmaps show quality varies across metrics, phrasing patterns, and models. Each heatmap shows one metric; rows are sub-properties and columns are models. Cells report mean scores, with darker green indicating higher quality.}
    \label{fig:rq3_quality_heatmap}
\end{figure*}

\subsection{RQ3: Energy-quality Trade-off Analysis}
\label{sec:rq3_energy_quality}

\begin{figure*}[t]
    \centering
    \includegraphics[
    width=\linewidth,
    height=0.48\textheight,
    keepaspectratio
]{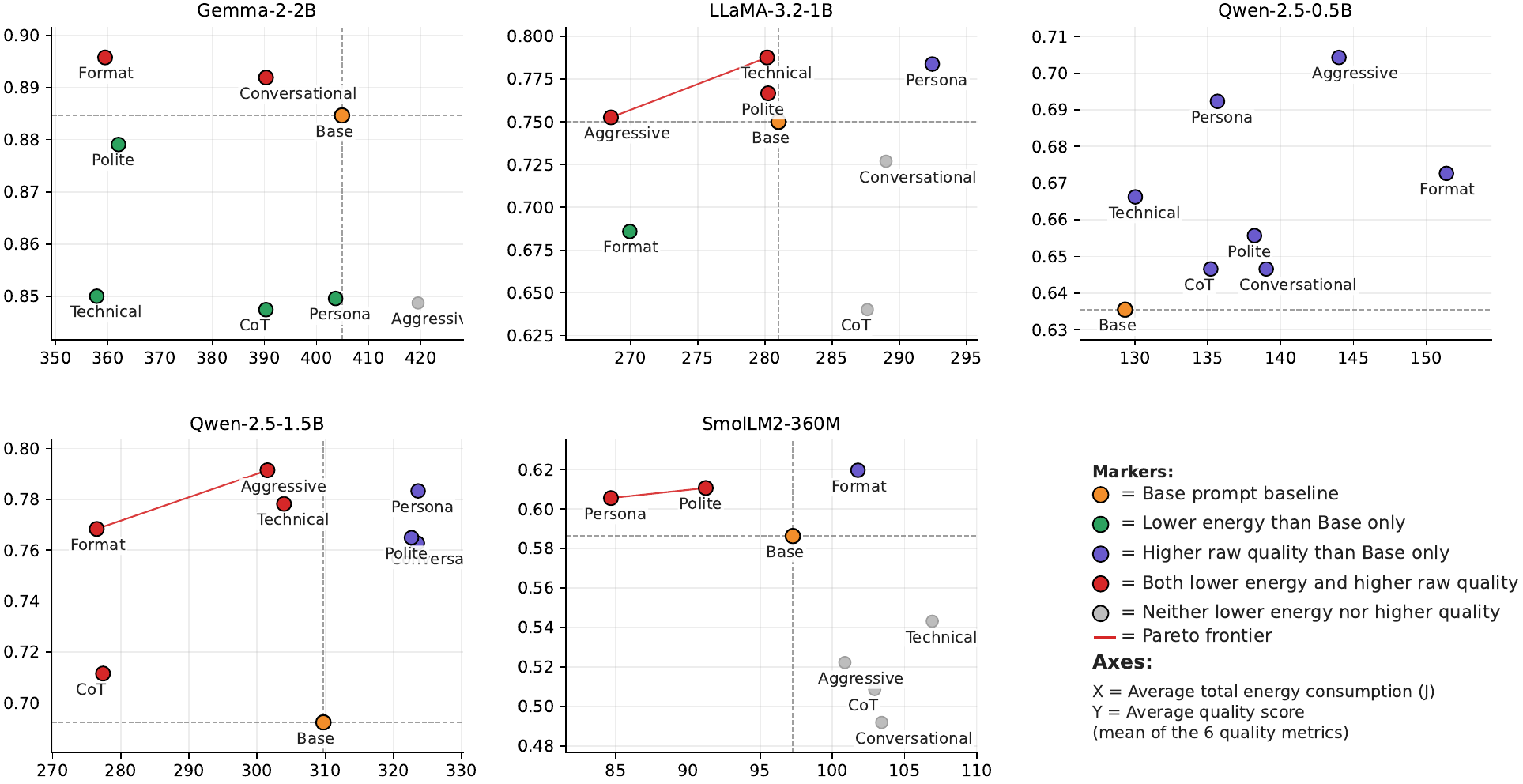}
    \caption{\textbf{Energy-quality trade-off and pareto frontier for phrasing pattern.}
    Each point represents a phrasing pattern sub-property evaluated under the same model weights and decoding configuration. Prompt sub-properties shift the energy-quality operating point in a model-dependent manner, revealing that stylistic variation can change both energy cost and response quality.}
    \label{fig:rq3_frontier}
\end{figure*}

For RQ3, we analyze the energy-quality trade-off by first examining response quality across phrasing pattern sub-properties and metrics, as shown in Figure~\ref{fig:rq3_quality_heatmap}. We then evaluate how each phrasing pattern shifts the energy-quality frontier under fixed model weights, as shown in Figure~\ref{fig:rq3_frontier}. Due to space constraints, we provide the remaining energy-quality results across models and metrics in Appendix~\ref{app:rq3_additional_results}.

\noindent\textit{Observation 1: Response quality varies across models and metrics.}
Figure~\ref{fig:rq3_quality_heatmap} shows that response quality is strongly model dependent. Larger or more capable models generally achieve higher scores across relevance, correctness, coherence, completeness, instruction adherence, and internal consistency, whereas smaller models show lower and less stable quality. This indicates that phrasing patterns interact with model capacity rather than producing uniform quality shifts.

\noindent\textit{Observation 2: Phrasing patterns affect quality dimensions unevenly.}
Phrasing patterns show relatively stable performance on relevance, coherence, and internal consistency, but larger variation on correctness, completeness, and instruction adherence. This indicates that stylistic changes can preserve the apparent fluency and alignment of a response, while improvements in task correctness and completeness remain less consistent.

\noindent \textit{Observation 3: Phrasing patterns reshape the energy-quality frontier.}
Figure~\ref{fig:rq3_frontier} shows that phrasing patterns can substantially alter each model's energy-quality profile. Red markers denote sub-properties achieving both lower energy and higher quality than Base. For LLaMA-3.2-1B, Qwen-2.5-1.5B, and SmolLM2-360M, these lie on the Pareto frontier, showing that prompt selection improves both sides of the trade-off under fixed weights and decoding settings. For Gemma-2-2B, the format and conversational prompts both improve on Base along both axes; the format prompt attains the highest quality and is the better operating point, though the technical prompt reaches slightly lower energy.

\noindent \textit{Observation 4: Energy-quality trade-offs are model-dependent.}
The energy-quality effect of a phrasing pattern does not transfer uniformly across models. A sub-property that improves the energy-quality balance for one model may lead to a different trade-off profile for another. This model dependence suggests that prompt phrasing style selection should be model-aware, rather than assumed to generalize uniformly across models.

\begin{tcolorbox}[
    colback=green!3!white,
    colframe=teal!80!black,
    boxrule=0.7pt,
    arc=1mm,
    left=1mm,
    right=1mm,
    top=1mm,
    bottom=1mm
]
\textit{\textbf{Finding.}} Phrasing patterns can substantially shift the energy-quality frontier, enabling lower energy consumption and higher response quality, while efficiency gains are model-dependent. This suggests that energy-efficient prompting is not governed by a uniform optimal phrasing style, but requires highly specific, model-aware phrasing optimization strategies.
\end{tcolorbox}

\section{Related Work}
\label{sec:related}

\textbf{Prompt categorization and linguistic analysis.}
Prior work has extensively studied prompt properties from linguistic and cognitive perspectives, including phrasing patterns \cite{NEURIPS2020_1457c0d6,10.1145/3411763.3451760,pmlr-v139-zhao21c}, syntactic and discourse structures \cite{schick-schutze-2021-exploiting,10.1162/tacl_a_00324,min-etal-2022-rethinking,liu-etal-2022-makes}, reasoning requirements \cite{NEURIPS2022_9d560961,NEURIPS2022_8bb0d291,wang2023selfconsistency}, and cognitive complexity \cite{fu2023complexitybased}. Surveys and taxonomies further systematize these properties along functional and cognitive axes \cite{10.1145/3560815,vatsal2024survey,liu2026comprehensive}, while recent frameworks model prompt structure to support interpretability and prompt engineering \cite{jeoung-etal-2026-promptprism}. However, this line of work focuses on output quality and robustness, with little attention to how prompt properties affect system-level behavior such as on-device energy consumption during inference.

\noindent \textbf{Prompt datasets and benchmarks.}
Large prompt and instruction datasets, such as Natural Instructions \cite{mishra-etal-2022-cross}, Super-NaturalInstructions \cite{wang-etal-2022-super}, and the Flan Collection \cite{pmlr-v202-longpre23a}, have been widely used for instruction tuning and evaluation. Tools like PromptSource further standardize prompt templates across tasks \cite{bach-etal-2022-promptsource}. While these resources enable broad coverage and performance-driven evaluation, prompts often vary simultaneously along multiple dimensions (e.g., semantics, length, and reasoning), making them unsuitable for isolating the impact of specific prompt properties on model or system behavior.

\noindent \textbf{On-device LLM systems and energy profiling.}
Recent studies benchmark latency, memory, and throughput of LLM inference on mobile and edge devices \cite{10.1145/3662006.3662059,murthy2024mobileaibench} and propose hardware-aware inference optimizations \cite{10.1145/3680207.3765259,xue2024powerinfer}. Energy profiling efforts typically analyze model architectures, runtimes, or prompt-level workloads on server-class hardware \cite{husom2024price,caravaca2025prompts}. Consequently, existing work provides limited insight into how prompt properties themselves influence on-device energy consumption under controlled conditions.

In contrast, we study prompt properties through direct measurements of on-device energy consumption, linking linguistic prompt analysis with system-level energy profiling. Our analysis is measurement-based and characterizes system-level energy behavior rather than isolating low-level execution mechanisms.

\section{Conclusion}
This paper studies how prompt properties shape the energy behavior and response quality of on-device LLM inference. Through a broad empirical study covering models, devices, and prompt variants, we show that prompt-level variants can substantially affect mobile inference efficiency. In particular, cognitively demanding prompts tend to increase decode energy cost to process each token, while surface-level phrasing patterns can change token usage, phase-level energy burden, and response quality without altering the underlying task. Our analysis further shows that these effects are not uniform: different prompts can shift cost between prefill and decode phases, and the resulting energy-quality trade-offs vary across models and metrics.

These findings suggest that prompt design is not only a usability or quality concern, but also a practical optimization lever for energy-aware LLM deployment. Rather than treating energy consumption as a fixed property of a model, future systems should account for how prompts shape generation behavior and inference cost, especially in mobile and resource-constrained settings. Future work will extend this analysis to a broader prompt taxonomy and more diverse hardware platforms.

\section*{Limitations}
\textbf{Limited prompt property coverage.} Our study primarily focuses on two prompt properties, cognitive load and phrasing pattern. However, the design space of prompts is vast, encompassing numerous other semantic and structural properties (e.g., emotional valence, stylistic constraints, or adversarial framings) that remain unexplored in our energy profiling. Future research incorporating a broader taxonomy of prompt properties would provide a more complete view of how diverse linguistic variables influence on-device energy consumption.

\noindent \textbf{Limited control over prompt length.} Extraneous load is defined by redundant or irrelevant information and intrinsic load by explicit step-by-step guidance; a prompt cannot carry these without becoming longer. Consequently, prompt length is not independently controlled, making it difficult to completely decouple the effects of cognitive load framing from length-related effects. Future work should construct length-matched prompt variants to better isolate these factors.

\noindent \textbf{Evaluation methodology limitations.} We use Gemini-2.5-Pro as the
primary evaluator for prompt property scoring and response-quality
assessment. Although this helps maintain internal consistency, relying
on a single LLM-as-a-judge may introduce model-specific biases in
scoring. In addition, embedding-based similarity and DeepEval metrics
provide automatic proxies for response quality, but may not fully
capture the nuanced utility perceived by human users. To strengthen evaluation reliability, we manually verify randomly sampled subsets of both prompt variation and response quality evaluations. Automated judgments that do not pass human review are replaced with human judgments.

\noindent \textbf{Limited hardware coverage and profiling granularity.} Our energy measurements are conducted on a limited set of mobile SoC architectures. Although these devices are representative of on-device inference settings, power profiles may differ across NPUs, GPUs, CPUs, memory systems, and runtime backends. In addition, our analysis focuses on end-to-end and phase-level energy behavior, rather than lower-level mechanisms such as KV-cache behavior, hardware counters, or operator-level traces. Future work should evaluate a wider range of mobile hardware and incorporate finer-grained profiling to explain hardware-specific energy variation.

\section*{Acknowledgments}
We thank the reviewers and the area chairs for their insightful comments. This research was supported by funds from Toyota Motor North America. In addition, this research was sponsored by the Army Research Laboratory and was accomplished under Cooperative Agreement Number W911NF-23-2-0224. The views and conclusions contained in this document are those of the authors and should not be interpreted as representing the official policies, either expressed or implied, of the Army Research Laboratory or the U.S. Government. The U.S. Government is authorized to reproduce and distribute reprints for Government purposes notwithstanding any copyright notation herein.


\bibliography{references}

\clearpage

\appendix
\label{sec:appendix}

\onecolumn

\section{Prompt Property Definitions}
\label{app:prompt_property_definitions}
\begin{table}[!htbp]
\centering
\footnotesize
\setlength{\tabcolsep}{4pt}
\renewcommand{\arraystretch}{1.2}
\begin{tabularx}{\columnwidth}{
  >{\raggedright\arraybackslash}p{0.30\columnwidth}
  >{\raggedright\arraybackslash}X
}
\toprule
\textbf{Sub-property} & \textbf{Definition} \\

\midrule
\textbf{Aggressive/Authoritative Tone} &
\textit{Prompts characterized by commanding or forceful language, often lacking politeness or courtesy.} \\
\textbf{Conversational Tone} &
\textit{Prompts that mimic natural human dialogue, often informal and friendly in nature.} \\
\textbf{Chain-of-Thought (CoT)} &
\textit{A prompting technique where the model is guided to generate intermediate reasoning steps before arriving at a final answer.} \\
\textbf{Formatting Differences} &
\textit{Variations in the structural presentation of prompts, such as the use of lists, bullet points, or different punctuation.} \\
\textbf{Persona-Based} &
\textit{Prompts that assign a specific role or identity to the model, such as “You are a helpful assistant.”} \\
\textbf{Polite Tone} &
\textit{Prompts that employ courteous language, including phrases like “please” and “thank you.”} \\
\textbf{Technical/Jargon-Heavy} &
\textit{Prompts that utilize domain-specific terminology or complex language.} \\
\bottomrule
\end{tabularx}
\caption{Definitions of phrasing pattern sub-properties, quoted from \citet{sotic2025university}.}
\label{tab:prompt_props_b}
\end{table}

\begin{table}[!htbp]
\centering
\small
\setlength{\tabcolsep}{4pt}
\renewcommand{\arraystretch}{1.2}
\begin{tabularx}{\columnwidth}{
  >{\raggedright\arraybackslash}p{0.30\columnwidth}
  >{\raggedright\arraybackslash}X
}
\toprule
\textbf{Sub-property} & \textbf{Definition} \\
\midrule
\textbf{Intrinsic Load} &
\textit{Prompts in this property explicitly guide models to break complex tasks into actionable steps aligned with LLM skills.} \\
\textbf{Extraneous Load} &
\textit{Prompts in this property contain unnecessary complexity with intricate language, redundant or irrelevant information, resulting in increased unnecessary load.} \\
\textbf{Germane Load} &
\textit{Prompts in this property explicitly engage models with their prior knowledge or deep working memory to integrate it with existing and new knowledge for problem-solving.} \\
\bottomrule
\end{tabularx}
\caption{Definitions of cognitive load sub-properties, quoted from ~\citet{long-etal-2025-makes}.}
\label{tab:prompt_props_a}
\end{table}

\section{Cognitive Load Prompt Generation Template}
\label{app:cogload_template}

\begin{prompttemplate}
\small\ttfamily

\vspace{0.6em}
PROPERTY = Cognitive Load

\vspace{0.6em}
DESCRIPTION = Cognitive Load Theory categorizes cognitive load into three types:
Intrinsic Load, Extraneous Load, and Germane Load. This property defines cognitive
load as a prompt attribute with three corresponding sub-properties.

\vspace{0.8em}
TASK INSTRUCTION = You are tasked with generating prompt variations that incorporate
cognitive load information based on the given prompt.

\vspace{0.8em}
SUB-PROPERTY DEFINITIONS:

\textgreater Intrinsic Load \\
Prompts explicitly guide the model to break complex tasks into actionable steps aligned
with LLM skills.

\textgreater Extraneous Load \\
Prompts contain unnecessary complexity with intricate language, redundant or irrelevant
information, resulting in increased unnecessary cognitive load.

\textgreater Germane Load \\
Prompts explicitly engage the model with its prior knowledge or deep working memory to
integrate existing and new knowledge for problem-solving.

\vspace{0.8em}
INPUT PROMPT FORMAT: \\
\textless Begin of the prompt\textgreater\ \{input\_prompt\}\ \textless End of the prompt\textgreater

\vspace{0.8em}
GENERATION REQUIREMENTS:

\textgreater Generate exactly one prompt variation for each cognitive load sub-property:
Intrinsic Load, Extraneous Load, and Germane Load.

\textgreater Preserve semantic intent:
Each variation must preserve the original semantic intent (the question and required
answer) while differing in the expression and organization of cognitive load information.
You may drop story details, background descriptions, or numbers that are not necessary
to determine the answer.

\textgreater Length consistency:
The three generated variations should be roughly comparable in length; however,
cognitive-load constraints take priority over strict length matching.
Do not artificially extend the Germane prompt to match the Extraneous prompt length.

\vspace{0.8em}
SUB-PROPERTY ISOLATION RULES:

1. Intrinsic Load variation:

- MUST include only step-breaking or task-structuring guidance.

- Use explicit step markers such as ``Step~1 / Step~2 / Step~3'' or
     ``First / Next / Then / Finally''.

- Do NOT add narrative details beyond the minimum required for solving the task.

2. Extraneous Load variation:

- MUST include only unnecessary complexity, redundancy, or irrelevant information.

- MUST NOT introduce step-by-step solution structure or meta-cognitive guidance.

3. Germane Load variation:

- MUST activate prior knowledge and conceptual reasoning.

- MUST NOT include explicit procedural or step-by-step guidance.

\vspace{0.8em}
CRITICAL CONSTRAINT CHECKLIST:

\textgreater For Extraneous Load Variation (Goal: High Extraneous, Low Intrinsic, Low Germane):

- Noise Injection: You MUST insert clearly irrelevant numerical data or distinct entities
     (e.g., unrelated times, prices, objects, names, weather).

- Distractors MUST NOT change the correct answer.

- Structure: Use complex and convoluted sentences to obscure key information.

- No Guidance: Do NOT use phrases such as ``First'', ``Next'', ``Think step-by-step'',
     or ``Use your knowledge of''.

\textgreater For Germane Load Variation (Goal: Low Intrinsic, Low Extraneous, High Germane):

- ONE-SENTENCE / ONE-MAIN-VERB RULE:
     The prompt MUST be exactly one sentence with one main directive verb.

- Explicit prior-knowledge cue is REQUIRED
     (e.g., ``Use your knowledge of the simple part-whole idea...'').

- Single simple concept ONLY:
     part-whole idea, part-part-whole idea, or difference idea.

- SEMANTIC SIMPLICITY:
     BANNED terms include ``Function'', ``Dependency'', ``Integration'',
     ``Framework'', ``Schema'', ``Mental model''.

- NUMBERS AND ROLES:
     Mention only quantities required to compute the answer.

- Target structure (soft template):
     ``Use your knowledge of the simple [concept] idea to find [goal] from
     the given numbers.''

\vspace{0.8em}
OUTPUT FORMAT (STRICT):

\textless Begin of response\textgreater

\{
  ``Base prompt'': ``\{input\_prompt\}'',
  
  ``Intrinsic load prompt'': ``'',
  
  ``Extraneous load prompt'': ``'',
  
  ``Germane load prompt'': ``''
\}

\textless End of response\textgreater

\vspace{0.8em}
Any deviation from the required format or violation of sub-property isolation is
strictly prohibited. No additional explanation or commentary is allowed outside the
JSON object.
\end{prompttemplate}

\section{Prompt Variation Evaluator Template}
\label{app:prompt_evaluator}

\begin{prompttemplate}
\small\ttfamily

\vspace{0.6em}
COG\_FORMAT = ``\{`Intrinsic load': 1-10, `Extraneous load': 1-10, `Germane load': 1-10\}''

\vspace{0.6em}
COG\_JUDGING\_PROMPT = f"""
You are a highly experienced judge tasked with evaluating a prompt on criteria.

The prompt given to you is provided below:
<begin of the prompt> \{input\_prompt\} <end of the prompt>

Your task is to evaluate the above prompt on the following criteria on a scale of 1-10:

- Intrinsic load: This evaluates the prompts in explicitly guiding models to break complex tasks into actionable steps aligned with LM skills.

- Extraneous load: The extent to which prompts exclude irrelevant materials to reduce unnecessary load.

- Germane load: The degree to which prompts explicitly engage models with their prior knowledge or deep working memory (e.g., ``ask itself'') to integrate it with existing and new knowledge for problem-solving.

The scoring system is provided below:

> Intrinsic load:

- 1-2 (Poor): The prompt provides little to no guidance on breaking down the task. It is overly vague, abstract, or assumes the model can handle complexity without guidance.

- 3-4 (Below Average): The prompt provides minimal guidance but fails to clearly break the task into actionable steps. The model is left to infer most of the process.

- 5-6 (Average): The prompt partially breaks down the task but lacks clarity or completeness in defining actionable steps. Some guidance is present, but it is inconsistent or incomplete.

- 7-8 (Good): The prompt effectively breaks the task into clear, actionable steps. It aligns well with the model’s skills but may lack some nuance or optimization.

- 9-10 (Excellent): The prompt perfectly breaks the task into logical, actionable steps. It is highly aligned with the model’s capabilities and ensures clarity and efficiency in execution.

> Extraneous load:

- 1-2 (Poor): The prompt is perfectly concise and excludes all irrelevant materials. It is optimized to reduce extraneous load to the bare minimum.

- 3-4 (Below Average): The prompt is concise and mostly free of irrelevant information. It minimizes extraneous load effectively, with only minor distractions.

- 5-6 (Average): The prompt includes some unnecessary details but generally stays focused on the task. The extraneous load is moderate but not overly detrimental.

- 7-8 (Good): The prompt contains some irrelevant information, but the core task is still somewhat discernible. The extraneous load is noticeable and distracting.

- 9-10 (Excellent): The prompt includes excessive irrelevant information, making it difficult for the model to focus on the core task. It is cluttered or overly verbose.

> Germane load:

- 1-2 (Poor): The prompt does not engage the model’s prior knowledge or working memory. It provides no cues or instructions to leverage existing knowledge.

- 3-4 (Below Average): The prompt makes minimal attempts to engage prior knowledge but does so ineffectively or inconsistently. The model is left to infer connections on its own.

- 5-6 (Average): The prompt partially engages the model’s prior knowledge but lacks depth or clarity in integrating it with new information. The engagement is superficial.

- 7-8 (Good): The prompt effectively engages the model’s prior knowledge and encourages integration with new information. It provides clear cues or instructions for leveraging existing knowledge.

- 9-10 (Excellent): The prompt perfectly engages the model’s prior knowledge and deep working memory. It explicitly guides the model to integrate existing and new knowledge for optimal problem-solving.

Your evaluations must focus on explicit instructions rather than implicit instructions.

For example, if the prompt does not say ``Reflect on your prior knowledge'' then you should not assume that the prompt is effective in encouraging germane load.

Begin your evaluation by providing a short explanation for each. Be as objective, thorough, and constructive as possible.

After providing your explanation, please rate the response on all the criteria on a scale of 1 to 10 by strictly following this format:
<begin of explanation> ... <end of explanation>
<begin of ratings> \{COG\_FORMAT\} <end of ratings>
"""
\end{prompttemplate}

\section{Experimental Setup Details}
\label{app:exp_setup}

\FloatBarrier
\begin{table*}[!htbp]
\centering
\footnotesize
\setlength{\tabcolsep}{6pt}
\resizebox{\textwidth}{!}{%
\begin{tabular}{l|l|p{5.4cm}|p{4.0cm}}
\hline
\textbf{Category} &
\textbf{Item} &
\textbf{Configuration / Value} &
\textbf{Notes} \\
\hline

\multirow{2}{*}{\textbf{Devices}}
& Pixel~8~Pro
& CPU: 1.32\,GHz / 1.57\,GHz / 2.04\,GHz \newline
  GPU: 810\,MHz
& Android on-device inference \\
& Pixel~7
& CPU: 1.32\,GHz / 1.49\,GHz / 1.58\,GHz \newline
  GPU: 810\,MHz
& Same OS configuration \\
\hline

\multirow{2}{*}{\textbf{Hardware Control}}
& CPU / GPU Frequency
& Fixed (no DVFS)
& Governors locked during inference \\
& Screen Brightness
& Minimum brightness
& Reduce display power noise \\
\hline

\multirow{5}{*}{\textbf{Models}}
& Llama-3.2-1B-Instruct
& 1B params, q4f16\_1, MLC
& Instruction-tuned \\
& Qwen-2.5-0.5B-Instruct
& 0.5B params, q0f16, MLC
& Lightweight model \\
& Qwen-2.5-1.5B-Instruct
& 1.5B params, q4f16\_1, MLC
& Medium-scale model \\
& Gemma-2-2b-it
& 2B params, q4f16\_0, MLC
& Google Gemma family \\
& SmolLM2-360M-Instruct
& 360M params, q4f16\_1, MLC
& Small-scale model \\
\hline

\multirow{3}{*}{\textbf{Prompt Setup}}
& Base Prompt Datasets
& SVAMP, BoolQ, AI2-ARC
& Used as semantic anchors \\
& Prompt Variations
& Property-controlled generation
& Cognitive load, phrasing patterns \\
& Prompt Count
& Three candidates per base prompt
& See Appendix~\ref{app:cogload_template} \\
\hline

\multirow{4}{*}{\textbf{Inference Settings}} 
& Number of Runs & Three runs per (device, model, prompt) & Averaged for stability \\
& Temperature    & $T = 0.6$ (fixed)                    & Same across all experiments \\
& Top-$p$         & $p = 0.8$ (fixed)                     & Same across all experiments \\
& Max Tokens     & Fixed across models                   & Prevent output-length bias \\
\hline

\textbf{Measurement}
& Energy Profiling
& On-device power measurement
& See Section~\ref{sec:profiling} \\
\hline
\end{tabular}
}
\caption{Experimental setup and controlled configurations for on-device prompt-energy evaluation. 
We conducted a carefully controlled study covering 2 major prompt properties, 10 sub-properties, 4 datasets, 404 prompts, 5 LLMs, 2 mobile devices (phrasing pattern prompts are profiled on both devices, cognitive load prompts are profiled on Pixel 8 Pro only), and 3 repeated runs per configuration, resulting in 7,620 total inference runs. 
To isolate the effect of prompt properties, CPU and GPU frequencies are fixed to disable DVFS, and all experiments use identical inference parameters across prompt variants.}
\label{tab:exp_setup}
\end{table*}
\FloatBarrier

\section{Phrasing Pattern Response Evaluation Metrics and Scoring Rubrics}
\label{app:metrics}

\begin{table}[t]
\centering
\small
\begin{tabular}{p{0.25\linewidth}p{0.68\linewidth}}
\toprule
\textbf{Metric} & \textbf{Definition} \\
\midrule
Relevance &
Whether the response directly addresses the user's request and remains aligned with the prompt's intent and scope, without extraneous or tangential content. \\

Correctness &
Whether the response avoids factual errors or hallucinated claims unsupported by the prompt or commonly accepted knowledge. \\

Coherence &
Whether the response is logically organized, structurally well formed, and semantically continuous, without abrupt topic shifts, disorganization, or repetitive generation. \\

Completeness &
Whether the response covers all material components of the request, including key sub-questions, required elements, or expected deliverables. \\

Instruction Adherence &
Whether the response follows explicit and implicit prompt constraints, including format, scope, quantity, style, and target audience. \\

Internal Consistency &
Whether the response is free of internal contradictions, self-negating statements, or mutually inconsistent claims. \\
\bottomrule
\end{tabular}
\caption{Definitions of the six response-quality metrics used to evaluate linguistic and functional competence in open-ended phrasing tasks.}
\label{tab:metric-definitions}
\end{table}

\begin{table*}[h]
    \centering
    \small
    \renewcommand{\arraystretch}{1.4}
    \caption{Scoring Rubric for \textbf{Relevance} and \textbf{Correctness}}
    \label{tab:rubric_rc}
    
    \begin{tabularx}{\textwidth}{@{} c | X | X @{}}
        \toprule
        \textbf{Score} & \multicolumn{1}{c|}{\textbf{Relevance}} & \multicolumn{1}{c}{\textbf{Correctness}} \\ 
        \midrule
        
        \textbf{1.00} 
        & \textbf{Completely Relevant.} Entirely focused on the user's prompt. Every sentence contributes to answering the specific question. 
        & \textbf{No Errors.} All assertions and claims are factually correct and verifiable based on general knowledge. \\ 
        \cmidrule(lr){1-3}
        
        \textbf{0.75} 
        & \textbf{Mostly Relevant.} Core answer is relevant, but includes slight divergence, tangential examples, or minor unnecessary elaboration. 
        & \textbf{Minor Inaccuracy.} Core answer is correct, but contains a trivial error (e.g., slightly off date) that does not mislead the user. \\ 
        \cmidrule(lr){1-3}
        
        \textbf{0.50} 
        & \textbf{Mixed Relevance.} Addresses the prompt but drifts significantly into unrelated topics for a large portion of the text. 
        & \textbf{Mixed Accuracy.} Contains a mix of correct and incorrect statements. A significant claim is factually wrong. \\ 
        \cmidrule(lr){1-3}
        
        \textbf{0.25} 
        & \textbf{Mostly Irrelevant.} Acknowledges the topic but fails to address the specific question; high noise-to-signal ratio. 
        & \textbf{Major Inaccuracy.} Primary conclusion is factually incorrect, though some minor supporting details might be right. \\ 
        \cmidrule(lr){1-3}
        
        \textbf{0.00} 
        & \textbf{Completely Irrelevant.} Unrelated to the prompt (e.g., hallucinated question, gibberish). 
        & \textbf{Complete Hallucination.} Fundamentally wrong, fabricated, or contradicts well-established facts entirely. \\ 
        \bottomrule
    \end{tabularx}
\end{table*}

\begin{table*}[h]
    \centering
    \small
    \renewcommand{\arraystretch}{1.4}
    \caption{Scoring Rubric for \textbf{Coherence} and \textbf{Completeness}}
    \label{tab:rubric_cc}
    
    \begin{tabularx}{\textwidth}{@{} c | X | X @{}}
        \toprule
        \textbf{Score} & \multicolumn{1}{c|}{\textbf{Coherence}} & \multicolumn{1}{c}{\textbf{Completeness}} \\ 
        \midrule
        
        \textbf{1.00} 
        & \textbf{Perfectly Coherent.} Flows logically with clear structure. Transitions are smooth and natural. 
        & \textbf{Fully Complete.} Comprehensively addresses every aspect, including all sub-questions and implied requirements. \\ 
        \cmidrule(lr){1-3}
        
        \textbf{0.75} 
        & \textbf{Mostly Coherent.} Logic is sound, but contains minor awkward transitions or slightly confusing structures. 
        & \textbf{Mostly Complete.} Addresses main points but misses a minor detail, example, or nuance. \\ 
        \cmidrule(lr){1-3}
        
        \textbf{0.50} 
        & \textbf{Somewhat Disjointed.} Understandable but requires effort to follow. Text may jump between ideas. 
        & \textbf{Partial Completion.} Answers one part well but ignores another significant part (e.g., explains concept but omits examples). \\ 
        \cmidrule(lr){1-3}
        
        \textbf{0.25} 
        & \textbf{Incoherent.} Fragmented text. Sentences do not logically follow one another. Severe repetition occurs. 
        & \textbf{Minimal Coverage.} Touches on the topic but fails to provide the substance required for a functional answer. \\ 
        \cmidrule(lr){1-3}
        
        \textbf{0.00} 
        & \textbf{Unreadable.} ``Word salad,'' gibberish, or completely unstructured text. 
        & \textbf{Incomplete.} Fails to answer the core request entirely. \\ 
        \bottomrule
    \end{tabularx}
\end{table*}

\FloatBarrier
\noindent
\begin{minipage}{\textwidth}
    \small
    \renewcommand{\arraystretch}{1.4}
    \captionof{table}{Scoring Rubric for \textbf{Instruction Adherence} and \textbf{Internal Consistency}}
    \label{tab:rubric_ii}
    
    \begin{tabularx}{\textwidth}{@{} c | X | X @{}}
        \toprule
        \textbf{Score} & \multicolumn{1}{c|}{\textbf{Instruction Adherence}} & \multicolumn{1}{c}{\textbf{Internal Consistency}} \\ 
        \midrule
        
        \textbf{1.00} 
        & \textbf{Strict Adherence.} Follows every constraint, including negative constraints (``do not...'') and stylistic requirements. 
        & \textbf{Consistent.} Perfectly consistent. No logical contradictions exist within the text. \\ 
        \cmidrule(lr){1-3}
        
        \textbf{0.75} 
        & \textbf{Minor Deviation.} Follows main instructions but misses a minor formatting constraint or tone is slightly off. 
        & \textbf{Mostly Consistent.} Consistent, but may contain slight ambiguity interpretable as a contradiction. \\ 
        \cmidrule(lr){1-3}
        
        \textbf{0.50} 
        & \textbf{Partial Adherence.} Ignores a major constraint (e.g., wrong format) but adheres to others. 
        & \textbf{Contradictory.} Contains a clear contradiction on a minor point without context. \\ 
        \cmidrule(lr){1-3}
        
        \textbf{0.25} 
        & \textbf{Major Deviation.} Ignores most specific constraints, adhering only to the general topic. 
        & \textbf{Heavily Contradictory.} Contradicts its own main thesis or conclusion. \\ 
        \cmidrule(lr){1-3}
        
        \textbf{0.00} 
        & \textbf{Non-Adherence.} Completely ignores specific instructions (e.g., wrong language, wrong format). 
        & \textbf{Illogical.} Riddled with self-negating statements; impossible to determine the stance. \\ 
        \bottomrule
    \end{tabularx}
\end{minipage}

\section{Detailed Full Experimental Measurements}
\label{app:experimental_results}

\textbf{Note on per-token conventions.} Token counts and energy values in the following tables are arithmetic means across questions within each condition. In contrast, the per-token energy reported in Fig.~\ref{fig:rq1_per_token_energy} is first computed separately for each question and then averaged across questions, giving equal weight to every question. As a result, the two approaches use different weighting schemes: taking the ratio of the aggregate means in the table implicitly gives greater weight to questions with more tokens. For example, for Qwen-2.5-1.5B on BoolQ (Extraneous, decode), the equal-weight convention used in Fig.~\ref{fig:rq1_per_token_energy} yields a relative per-token energy of 1.29$\times$, while computing the ratio from the aggregate means in Table~\ref{tab:cognitive-load-boolq-overview} gives $(93.43/166) \div (25.56/56) = 1.23\times$. Both values are derived from the same underlying measurements and differ only in how questions are weighted. All per-token quantities reported in the text and Fig.~\ref{fig:rq1_per_token_energy} follow the equal-weight, per-question convention.

\input{tables/phrasing_pattern_overview_8p}
\input{tables/phrasing_pattern_overview_7}
\input{tables/cognitive_load_ai2_arc_overview_7}
\input{tables/cognitive_load_boolq_overview_7}
\input{tables/cognitive_load_svamp_overview_7}

\section{Additional Figures for RQ1}
\label{app:rq1_additional_results}

\vspace{0.3em}
\noindent
\begin{minipage}[t]{0.47\textwidth}
\textbf{Overview.}
This appendix provides additional per-token energy results supporting the analysis in Section~\ref{sec:rq1_per_token_energy}. These figures complement Figure~\ref{fig:rq1_per_token_energy} by covering additional datasets and device.
\end{minipage}
\hfill
\begin{minipage}[t]{0.47\textwidth}
Each figure reports prefill and decode per-token energy. The top row shows absolute values, while the bottom row normalizes each sub-property by the corresponding Base prompt.
\end{minipage}
\vspace{0.5em}

\begin{figure*}[!htbp]
    \centering
    \includegraphics[width=\linewidth]{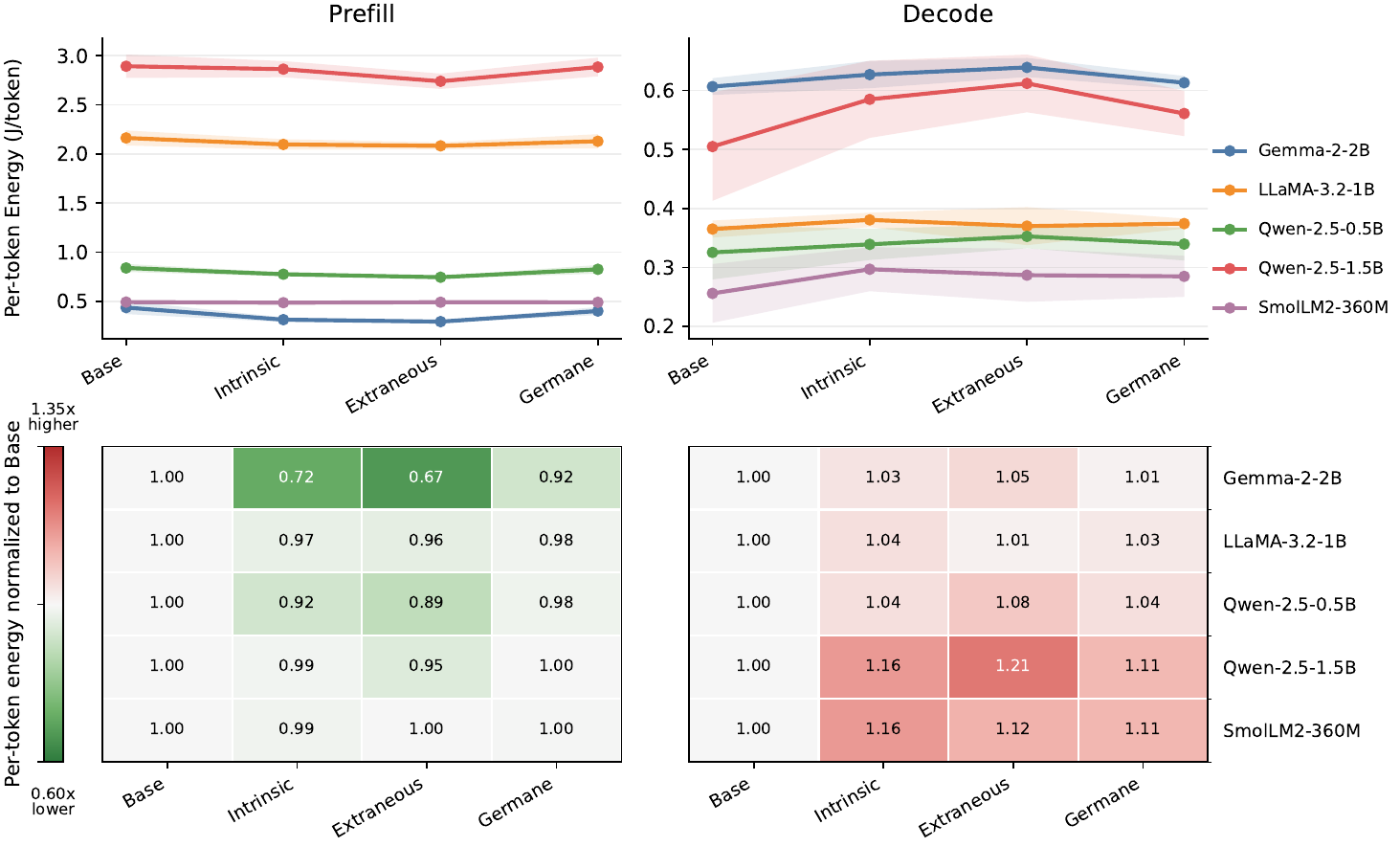}
    \caption{
    Additional per-token energy results for cognitive load on AI2-ARC on Pixel 8 Pro. 
    The figure reports prefill and decode per-token energy across intrinsic, extraneous, and germane load variants.}
    \label{fig:app_rq1_cognitive_ai2arc}
\end{figure*}

\begin{figure*}[!htbp]
    \centering
    \includegraphics[width=\linewidth]{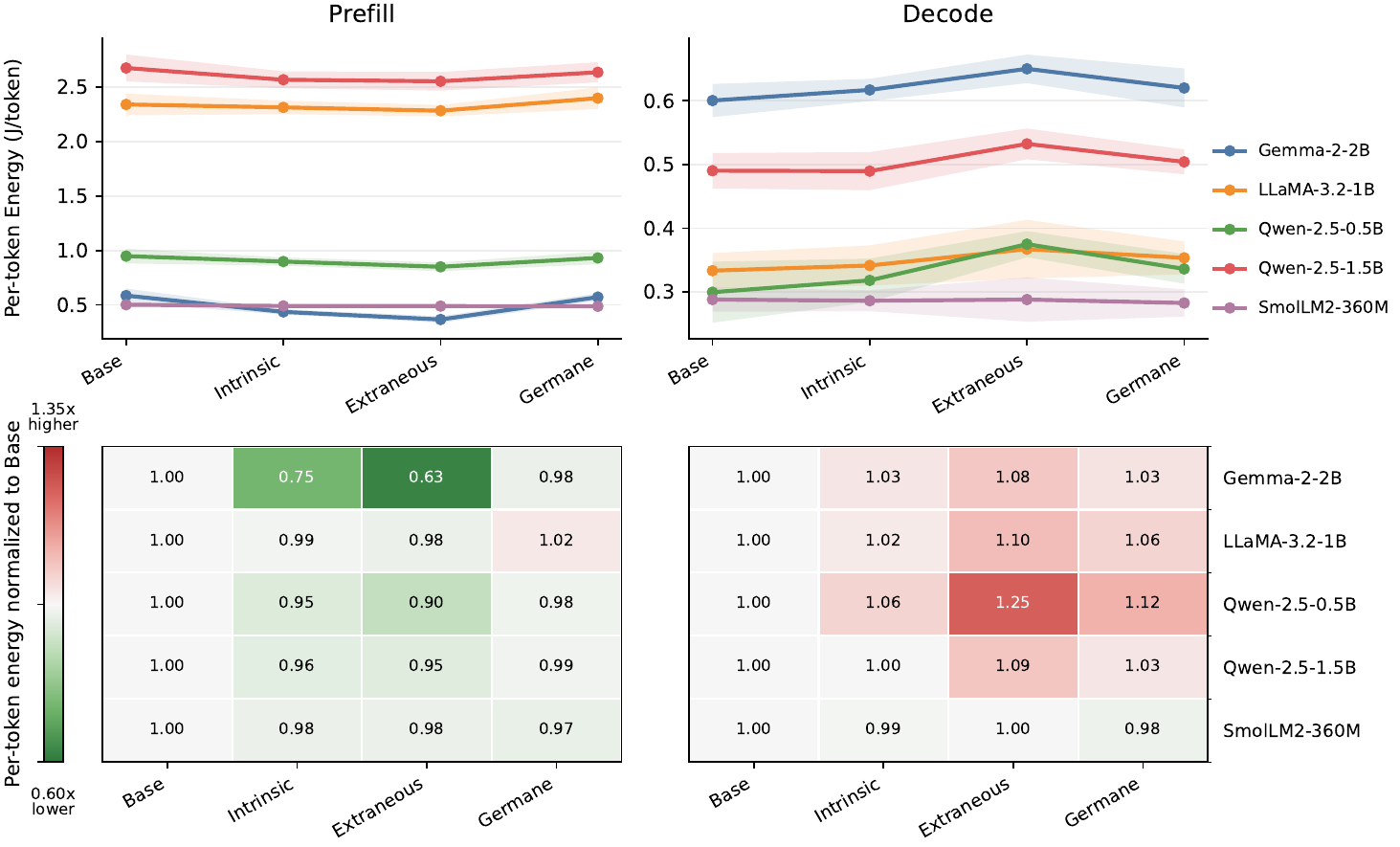}
    \caption{
    Additional per-token energy results for cognitive load on SVAMP on Pixel 8 Pro. The figure reports prefill and decode per-token energy across intrinsic, extraneous, and germane load variants.}
    \label{fig:app_rq1_cognitive_svamp}
\end{figure*}

\begin{figure*}[!htbp]
    \centering
    \includegraphics[width=\linewidth]{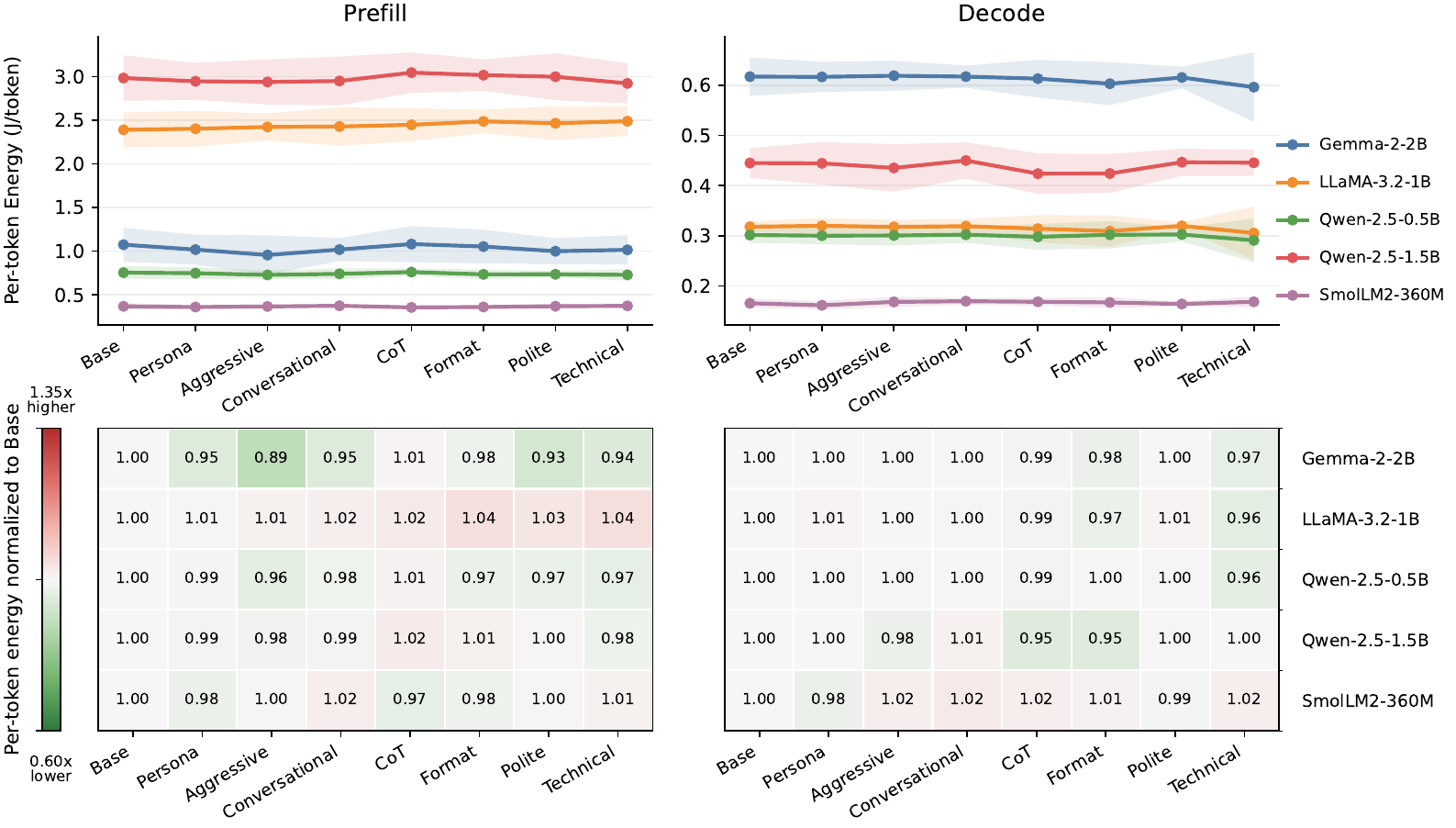}
    \caption{
    Additional per-token energy results for phrasing pattern  on Pixel 7. 
    The figure reports prefill and decode per-token energy across phrasing-pattern sub-properties.}
    \label{fig:app_rq1_phrasing}
\end{figure*}

\section{Additional Figures for RQ2}
\label{app:rq2_additional_results}

\vspace{0.3em}
\noindent
\begin{minipage}[t]{0.47\textwidth}
\textbf{Overview.}
This appendix provides additional token-usage and fixed-baseline burden results supporting the analysis in Section~\ref{sec:rq2_token_usage}. These figures complement Figure~\ref{fig:rq2_combined_tokenratio_phaseburden} by covering additional datasets and prompt-property settings. The top row reports total token ratio relative to Base.
\end{minipage}
\hfill
\begin{minipage}[t]{0.47\textwidth}
The bottom row reports fixed-baseline prefill and decode burden relative to Base. Bottom-row values are normalized by Base prompt/completion token counts, respectively, after subtracting the corresponding Base burden.
\end{minipage}
\vspace{0.5em}

\begin{figure*}[!htbp]
    \centering
    \includegraphics[width=\linewidth]{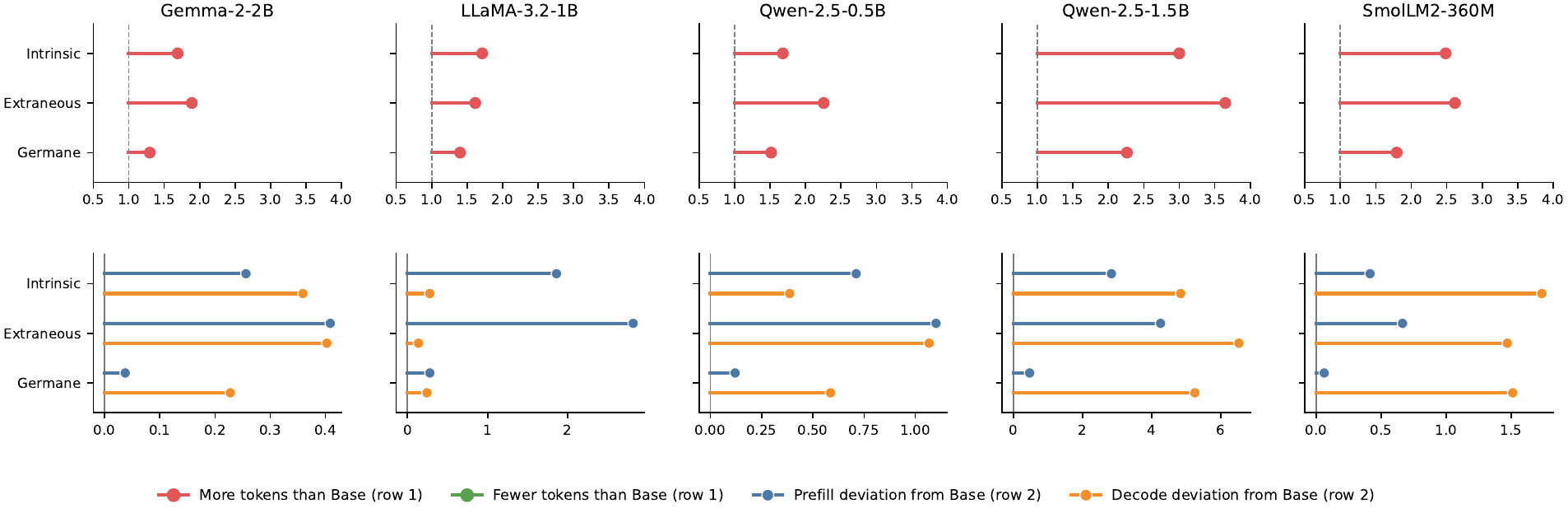}
    \caption{
    Additional token usage results for cognitive load on AI2-ARC on Pixel 8 Pro. 
    The top row reports total token ratio relative to Base across cognitive load variants. 
    The bottom row reports fixed-baseline prefill and decode burden relative to Base.
    }
    \label{fig:app_rq2_cognitive_ai2arc}
\end{figure*}

\begin{figure*}[!htbp]
    \centering
    \includegraphics[width=\linewidth]{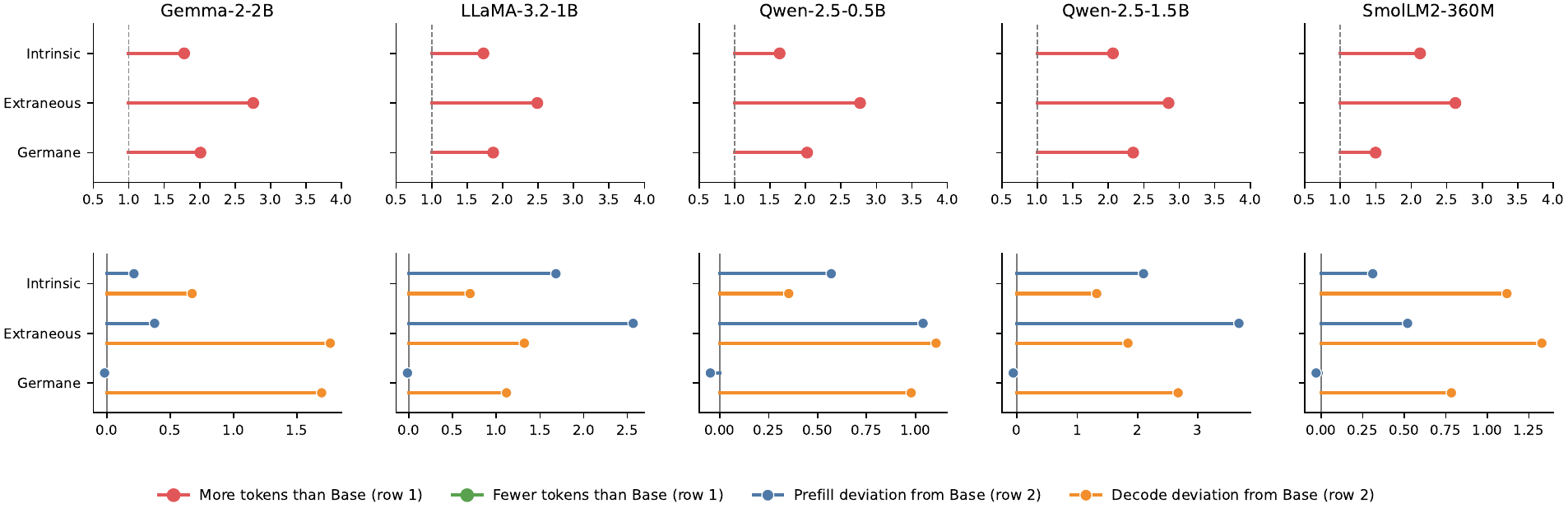}
    \caption{
    Additional token usage results for cognitive load on BoolQ on Pixel 8 Pro. 
    The top row reports total token ratio relative to Base across cognitive load variants. 
    The bottom row reports fixed-baseline prefill and decode burden relative to Base.
    }
    \label{fig:app_rq2_cognitive_boolq}
\end{figure*}

\begin{figure*}[!htbp]
    \centering
    \includegraphics[width=\linewidth]{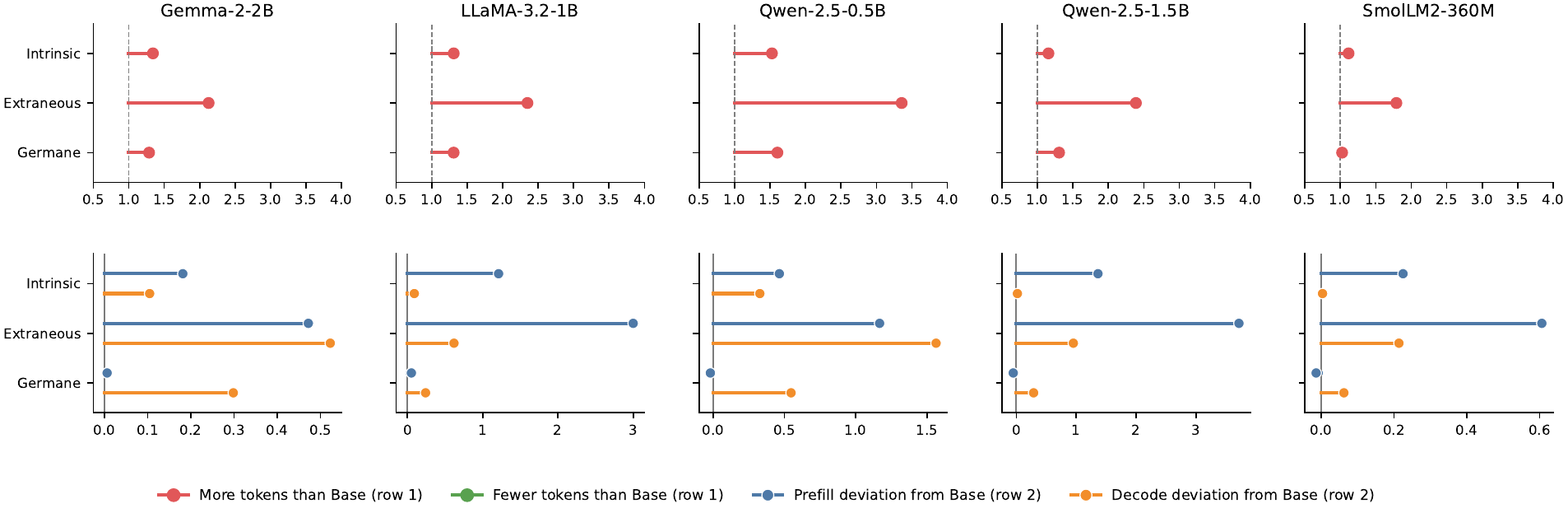}
    \caption{
    Additional token usage results for cognitive load on SVAMP. 
    The top row reports total token ratio relative to Base across cognitive load variants. 
    The bottom row reports fixed-baseline prefill and decode burden relative to Base.
    }
    \label{fig:app_rq2_cognitive_svamp}
\end{figure*}

\begin{figure*}[!htbp]
    \centering
    \includegraphics[width=\linewidth]{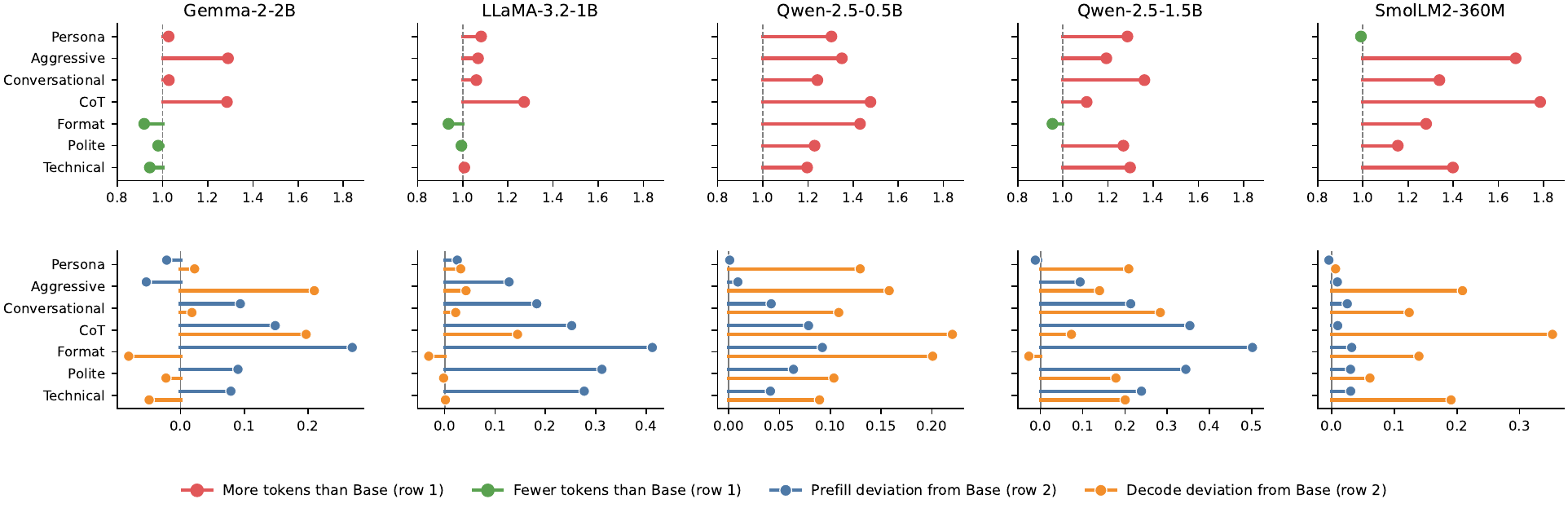}
    \caption{
    Additional token usage results for phrasing pattern  on Pixel 7. 
    The top row reports total token ratio relative to Base across phrasing-pattern sub-properties. The bottom row reports fixed-baseline prefill and decode burden relative to Base.
    }
    \label{fig:app_rq2_phrasing}
\end{figure*}

\section{Additional Figures for RQ3}
\label{app:rq3_additional_results}

\vspace{0.3em}
\noindent
\begin{minipage}[t]{0.47\textwidth}
\textbf{Overview.}
This appendix provides additional response-quality and energy-quality trade-off results supporting the analysis in Section~\ref{sec:rq3_energy_quality}.
\end{minipage}
\hfill
\begin{minipage}[t]{0.47\textwidth}
These figures complement the main-text analysis by covering cognitive-load accuracy results and additional trade-off settings.
\end{minipage}
\vspace{0.5em} 

\begin{figure*}[!htbp]
    \centering
    \includegraphics[width=\linewidth]{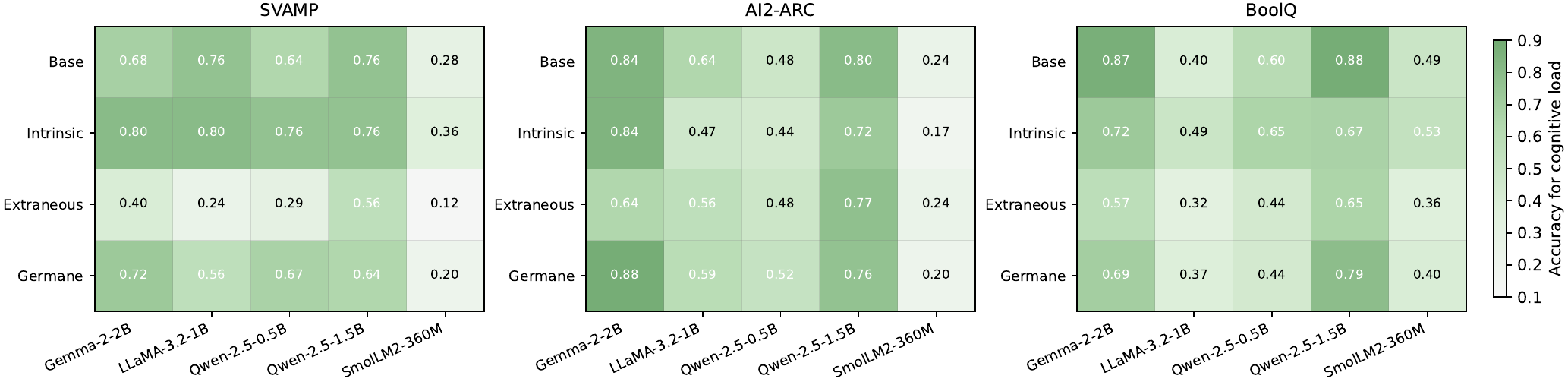}
    \caption{
    Additional RQ3 results for cognitive load response accuracy. 
    Each heatmap reports accuracy across cognitive load sub-properties, datasets, and models. 
    These results complement the response quality analysis in the main text by showing how intrinsic, extraneous, and germane load affect task correctness under objective ground-truth evaluation.
    }
    \label{fig:app_rq3_cognitive_accuracy_heatmaps}
\end{figure*}

\begin{figure*}[!htbp]
    \centering
    \includegraphics[width=\linewidth]{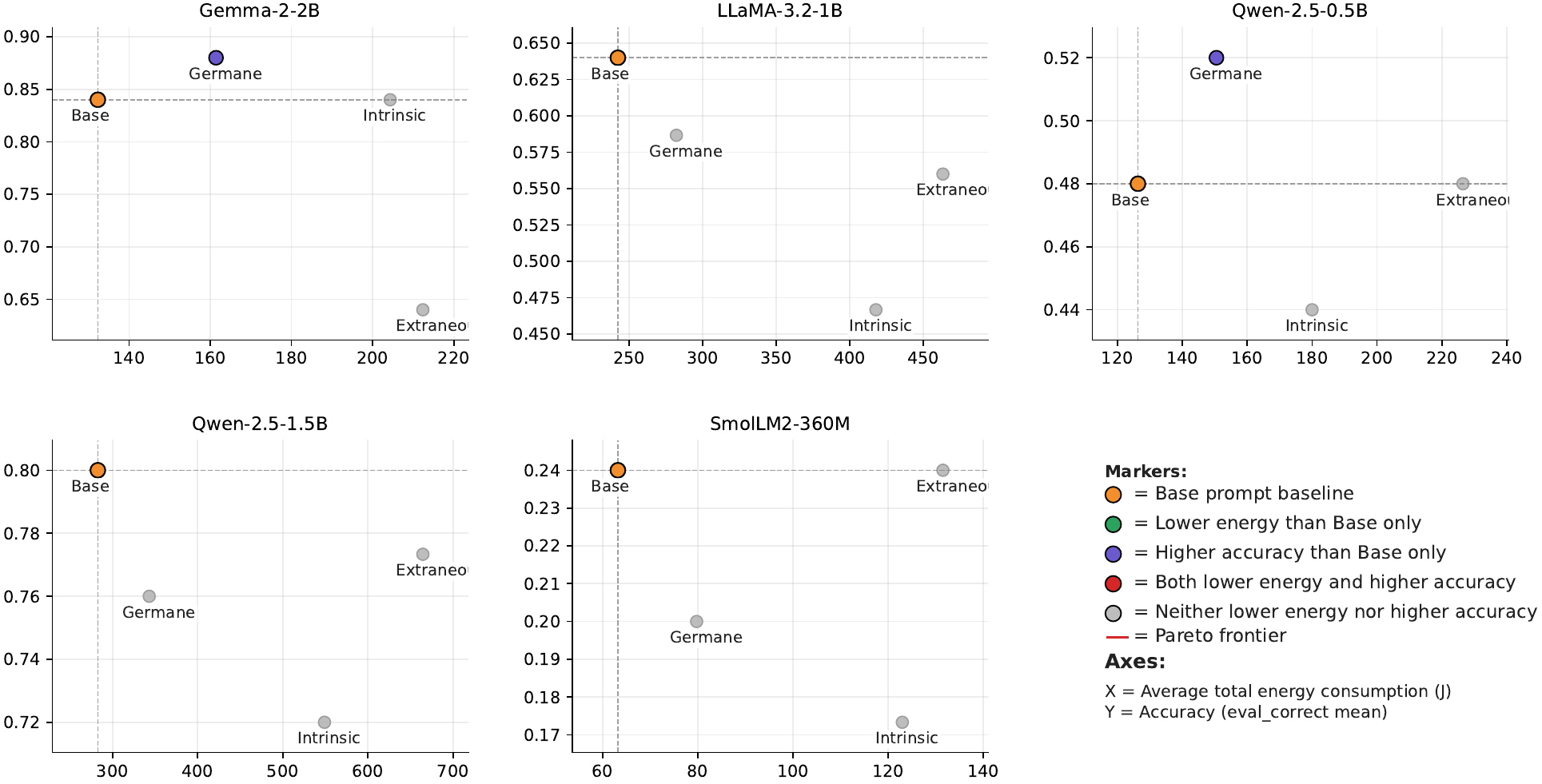}
    \caption{
    Additional RQ3 energy-accuracy trade-off results for cognitive load on AI2-ARC. 
    Each point represents a cognitive load sub-property under fixed model weights and decoding settings. 
    The x-axis reports average total energy consumption, and the y-axis reports accuracy.
    }
    \label{fig:app_rq3_cognitive_ai2arc_tradeoff}
\end{figure*}

\begin{figure*}[!htbp]
    \centering
    \includegraphics[width=\linewidth]{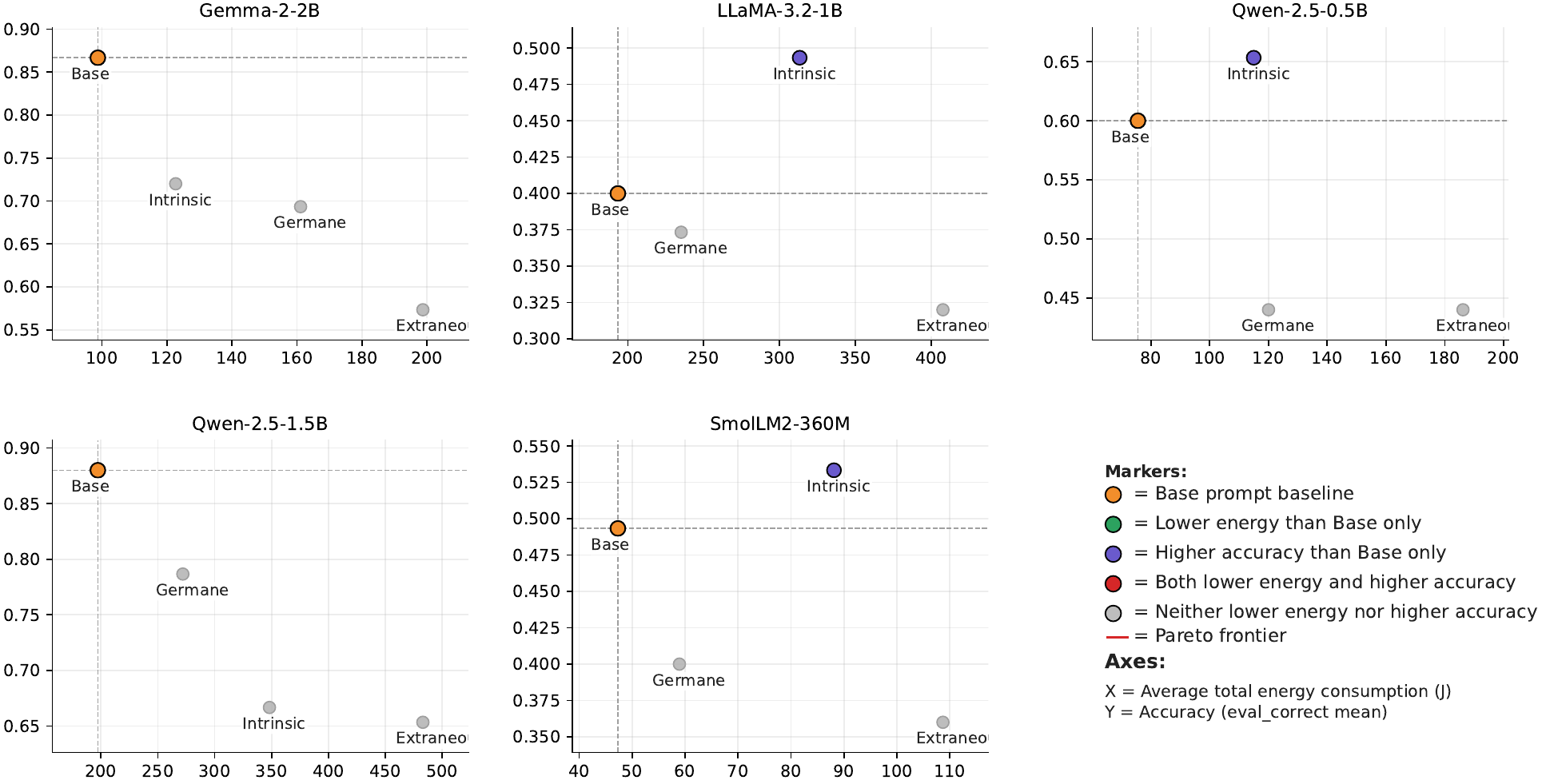}
    \caption{
    Additional RQ3 energy-accuracy trade-off results for cognitive load on BoolQ. 
    Each point represents a cognitive load sub-property under fixed model weights and decoding settings. 
    The x-axis reports average total energy consumption, and the y-axis reports accuracy.
    }
    \label{fig:app_rq3_cognitive_boolq_tradeoff}
\end{figure*}

\begin{figure*}[!htbp]
    \centering
    \includegraphics[width=\linewidth]{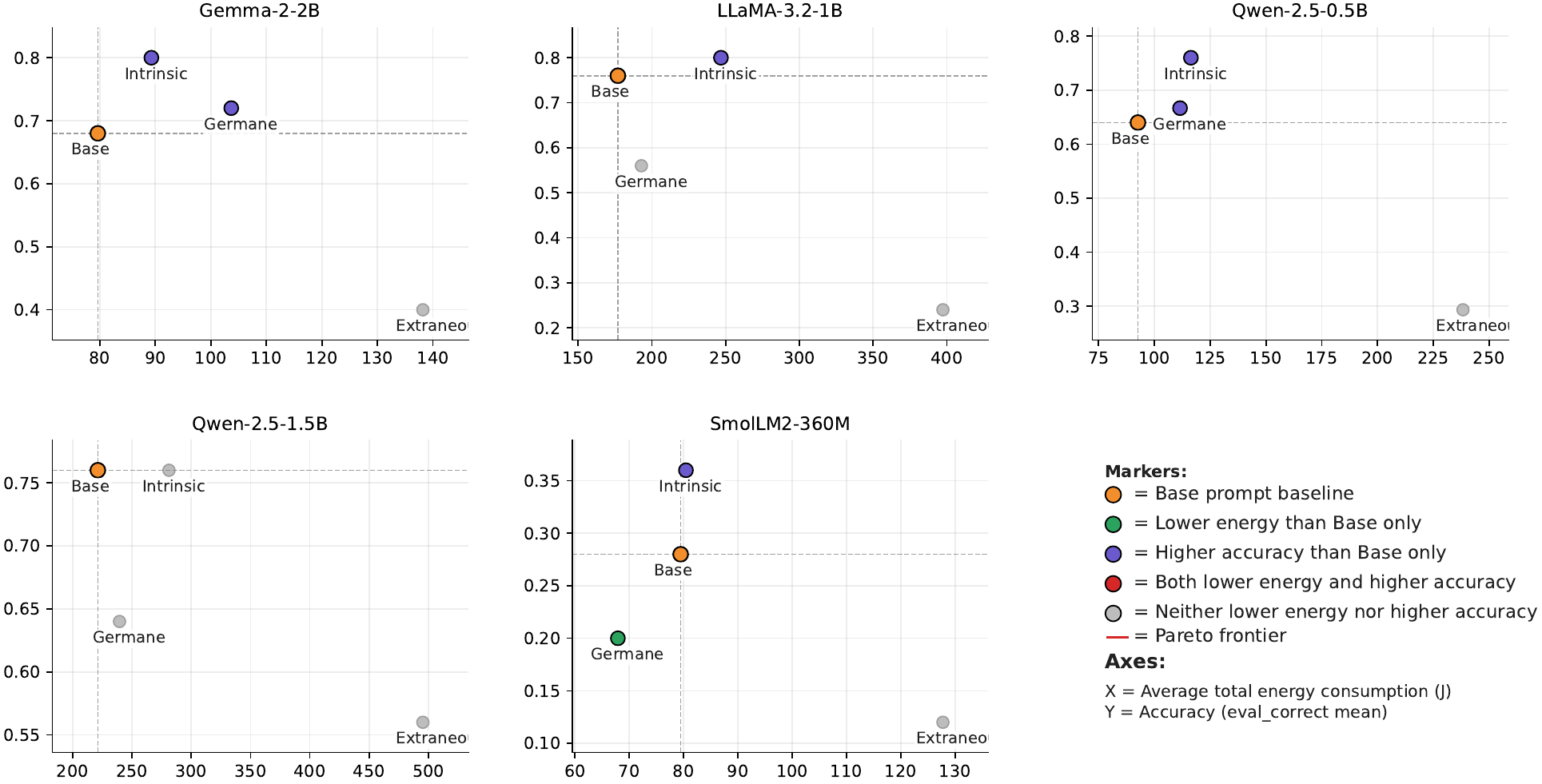}
    \caption{
    Additional RQ3 energy-accuracy trade-off results for cognitive load on SVAMP. 
    Each point represents a cognitive load sub-property under fixed model weights and decoding settings. 
    The x-axis reports average total energy consumption, and the y-axis reports accuracy.
    }
    \label{fig:app_rq3_cognitive_svamp_tradeoff}
\end{figure*}

\begin{figure*}[!htbp]
    \centering
    \includegraphics[width=\linewidth]{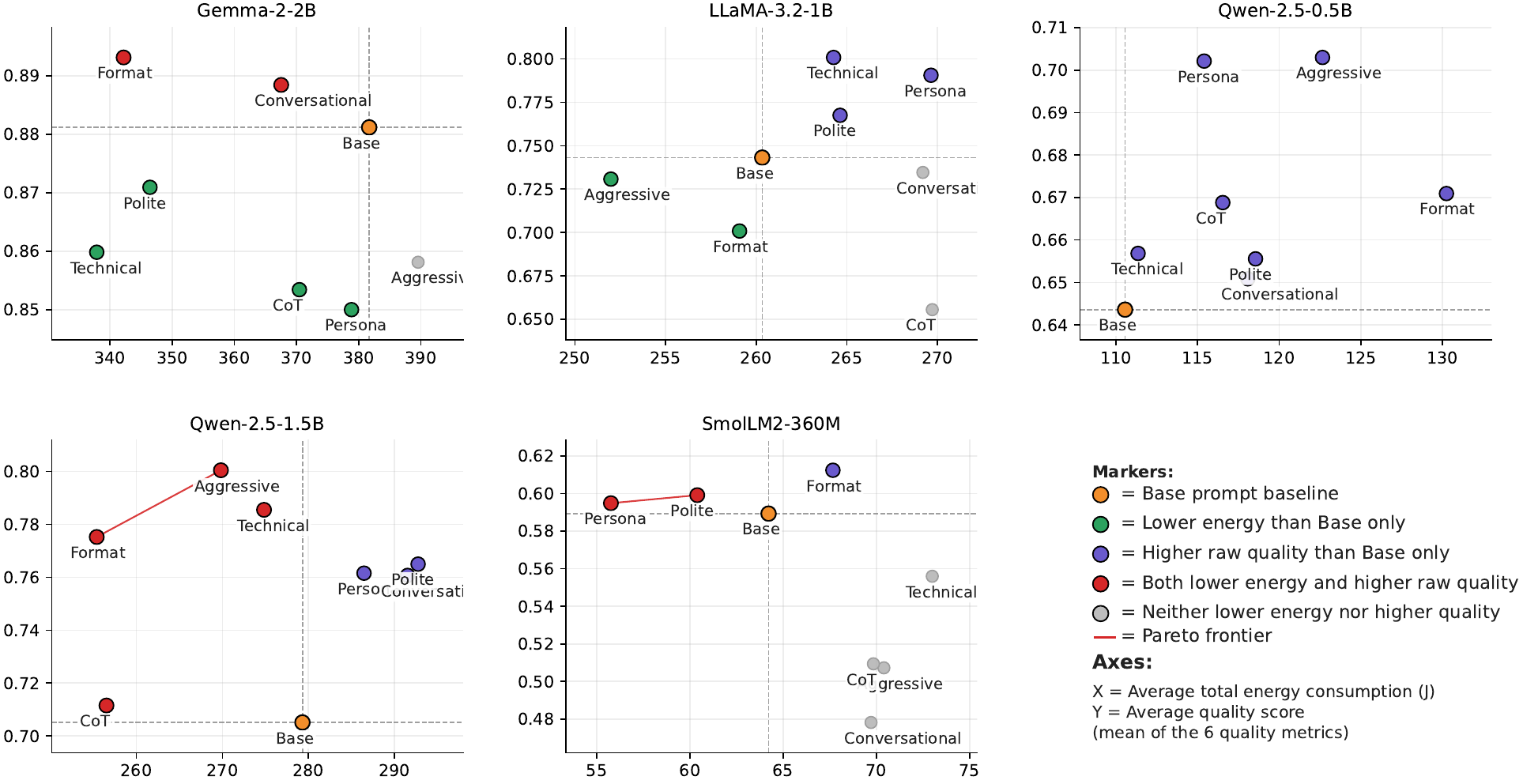}
    \caption{
    Additional RQ3 energy-quality trade-off results for phrasing on Pixel 7. Each point represents a phrasing pattern sub-property under fixed model weights and decoding settings. The x-axis reports average total energy consumption, and the y-axis reports the averaged response quality score across the six reference-free evaluation dimensions.}
    \label{fig:app_rq3_phrasing_tradeoff}
\end{figure*}

\section{License}
\label{License}
\vspace{0.3em}
\noindent
\begin{minipage}[t]{0.47\textwidth}
\textbf{Datasets.}
The datasets used in this study are governed by their respective licenses and data-use terms. SVAMP is released under the \textit{MIT License}; BoolQ under \textit{Creative Commons Attribution-ShareAlike 3.0 Unported (CC BY-SA 3.0)}; AI2-ARC under \textit{Creative Commons Attribution-ShareAlike 4.0 International (CC BY-SA 4.0)}; and CLEF 2025 data under the applicable \textit{Data Use Agreements}.
\end{minipage}
\hfill
\begin{minipage}[t]{0.47\textwidth}
\textbf{Software, models, and hardware.}
DeepEval is used under the \textit{Apache-2.0}, and MLC-LLM under the \textit{Apache-2.0}. The open-source models Llama-3.2, Qwen-2.5, Gemma-2, and SmolLM2 are used under their respective community licenses. Gemini-2.5-Pro is accessed via the \textit{Google AI applicable terms}. Hardware profiling on Pixel 7 and Pixel 8 Pro follows standard consumer and developer terms.
\end{minipage}
\vspace{0.5em}

\end{document}

%% file: tables/phrasing_pattern_overview_8p.tex

\begin{table*}[p]
\centering
\rotatebox{90}{%
\begin{minipage}{0.92\textheight}
\centering
\captionof{table}{
Phrasing pattern summary across five LLMs on Pixel 8 pro. Energy columns are shaded green, where darker indicates lower energy, and quality columns are shaded blue, where darker indicates higher quality.
}

\label{tab:phrasing-pattern-8p-overview}

\scriptsize
\setlength{\tabcolsep}{2.2pt}
\renewcommand{\arraystretch}{0.9}

\resizebox{\linewidth}{!}{%
\begin{tabular}{llrrrrrrrrrrrrrrr}
\toprule
Model & Phrasing &
\makecell{Prompt\\Tok.} &
\makecell{Completion\\Tok.} &
\makecell{Prefill\\Power (W)} &
\makecell{Decode\\Power (W)} &
\makecell{Prefill\\Latency (s)} &
\makecell{Decode\\Latency (s)} &
\makecell{Total\\Latency (s)} &
\makecell{Prefill\\Energy (J)} &
\makecell{Decode\\Energy (J)} &
Rel. & Corr. & Coh. & Comp. &
\makecell{Instr.\\Adh.} &
\makecell{Internal\\Cons.} \\
\midrule
    \multirow{8}{*}{Gemma-2-2B} & Base & 41 & 538 & 4.01 & 6.09 & 5.85 & 63.01 & 68.86 & \cellcolor[HTML]{7FC97F}23.71 & \cellcolor[HTML]{AEDEA7}381.19 & \cellcolor[HTML]{77B5D9}0.91 & \cellcolor[HTML]{69ADD5}0.91 & \cellcolor[HTML]{85BCDC}0.89 & \cellcolor[HTML]{77B5D9}0.83 & \cellcolor[HTML]{69ADD5}0.82 & \cellcolor[HTML]{77B5D9}0.96 \\
     & Persona & 41 & 533 & 4.01 & 6.06 & 5.75 & 63.62 & 69.37 & \cellcolor[HTML]{72C375}23.31 & \cellcolor[HTML]{A3DA9D}380.36 & \cellcolor[HTML]{B7D4EA}0.87 & \cellcolor[HTML]{A0CBE2}0.88 & \cellcolor[HTML]{B7D4EA}0.85 & \cellcolor[HTML]{92C4DE}0.81 & \cellcolor[HTML]{A0CBE2}0.77 & \cellcolor[HTML]{ABD0E6}0.92 \\
     & Aggressive & 40 & 556 & 4.06 & 6.01 & 5.61 & 66.54 & 72.15 & \cellcolor[HTML]{63BC6E}22.93 & \cellcolor[HTML]{B8E3B2}396.55 & \cellcolor[HTML]{ABD0E6}0.88 & \cellcolor[HTML]{ABD0E6}0.85 & \cellcolor[HTML]{ABD0E6}0.86 & \cellcolor[HTML]{ABD0E6}0.80 & \cellcolor[HTML]{92C4DE}0.77 & \cellcolor[HTML]{A0CBE2}0.92 \\
     & Conversational & 42 & 519 & 4.04 & 6.02 & 5.82 & 61.11 & 66.93 & \cellcolor[HTML]{8BCF89}23.72 & \cellcolor[HTML]{97D492}366.61 & \cellcolor[HTML]{69ADD5}0.92 & \cellcolor[HTML]{5DA5D1}0.93 & \cellcolor[HTML]{77B5D9}0.89 & \cellcolor[HTML]{85BCDC}0.82 & \cellcolor[HTML]{77B5D9}0.81 & \cellcolor[HTML]{5DA5D1}0.98 \\
     & CoT & 44 & 518 & 4.01 & 6.02 & 6.02 & 60.57 & 66.59 & \cellcolor[HTML]{AEDEA7}24.32 & \cellcolor[HTML]{8BCF89}365.98 & \cellcolor[HTML]{A0CBE2}0.90 & \cellcolor[HTML]{92C4DE}0.88 & \cellcolor[HTML]{A0CBE2}0.87 & \cellcolor[HTML]{B7D4EA}0.79 & \cellcolor[HTML]{B7D4EA}0.75 & \cellcolor[HTML]{B7D4EA}0.89 \\
     & Format & 45 & 476 & 4.03 & 6.00 & 6.06 & 55.49 & 61.55 & \cellcolor[HTML]{B8E3B2}24.54 & \cellcolor[HTML]{72C375}334.93 & \cellcolor[HTML]{5DA5D1}0.92 & \cellcolor[HTML]{77B5D9}0.89 & \cellcolor[HTML]{5DA5D1}0.90 & \cellcolor[HTML]{5DA5D1}0.87 & \cellcolor[HTML]{5DA5D1}0.83 & \cellcolor[HTML]{85BCDC}0.96 \\
     & Polite & 44 & 485 & 4.03 & 6.08 & 5.89 & 55.65 & 61.54 & \cellcolor[HTML]{A3DA9D}24.00 & \cellcolor[HTML]{7FC97F}338.03 & \cellcolor[HTML]{92C4DE}0.91 & \cellcolor[HTML]{92C4DE}0.88 & \cellcolor[HTML]{69ADD5}0.90 & \cellcolor[HTML]{69ADD5}0.84 & \cellcolor[HTML]{85BCDC}0.78 & \cellcolor[HTML]{69ADD5}0.97 \\
     & Technical & 43 & 473 & 4.03 & 5.93 & 5.85 & 55.49 & 61.34 & \cellcolor[HTML]{97D492}23.73 & \cellcolor[HTML]{63BC6E}334.13 & \cellcolor[HTML]{92C4DE}0.91 & \cellcolor[HTML]{B7D4EA}0.84 & \cellcolor[HTML]{92C4DE}0.88 & \cellcolor[HTML]{ABD0E6}0.80 & \cellcolor[HTML]{B7D4EA}0.75 & \cellcolor[HTML]{92C4DE}0.93 \\
    \midrule
    \multirow{8}{*}{LLaMA-3.2-1B} & Base & 56 & 380 & 3.57 & 6.36 & 34.99 & 24.45 & 59.45 & \cellcolor[HTML]{72C375}125.08 & \cellcolor[HTML]{A3DA9D}155.94 & \cellcolor[HTML]{85BCDC}0.81 & \cellcolor[HTML]{92C4DE}0.70 & \cellcolor[HTML]{77B5D9}0.82 & \cellcolor[HTML]{69ADD5}0.71 & \cellcolor[HTML]{92C4DE}0.66 & \cellcolor[HTML]{85BCDC}0.79 \\
     & Persona & 56 & 409 & 3.53 & 6.35 & 34.83 & 26.54 & 61.37 & \cellcolor[HTML]{63BC6E}123.59 & \cellcolor[HTML]{B8E3B2}168.87 & \cellcolor[HTML]{5DA5D1}0.84 & \cellcolor[HTML]{77B5D9}0.73 & \cellcolor[HTML]{69ADD5}0.83 & \cellcolor[HTML]{85BCDC}0.68 & \cellcolor[HTML]{5DA5D1}0.73 & \cellcolor[HTML]{5DA5D1}0.89 \\
     & Aggressive & 56 & 351 & 3.55 & 6.32 & 35.21 & 22.56 & 57.76 & \cellcolor[HTML]{7FC97F}125.68 & \cellcolor[HTML]{72C375}142.87 & \cellcolor[HTML]{92C4DE}0.80 & \cellcolor[HTML]{85BCDC}0.72 & \cellcolor[HTML]{85BCDC}0.82 & \cellcolor[HTML]{85BCDC}0.68 & \cellcolor[HTML]{A0CBE2}0.65 & \cellcolor[HTML]{69ADD5}0.84 \\
     & Conversational & 58 & 385 & 3.57 & 6.39 & 36.31 & 24.78 & 61.09 & \cellcolor[HTML]{8BCF89}130.57 & \cellcolor[HTML]{AEDEA7}158.45 & \cellcolor[HTML]{A0CBE2}0.79 & \cellcolor[HTML]{A0CBE2}0.68 & \cellcolor[HTML]{ABD0E6}0.77 & \cellcolor[HTML]{A0CBE2}0.67 & \cellcolor[HTML]{85BCDC}0.67 & \cellcolor[HTML]{B7D4EA}0.77 \\
     & CoT & 59 & 374 & 3.58 & 6.30 & 37.00 & 24.26 & 61.25 & \cellcolor[HTML]{A3DA9D}133.16 & \cellcolor[HTML]{97D492}154.47 & \cellcolor[HTML]{B7D4EA}0.64 & \cellcolor[HTML]{B7D4EA}0.62 & \cellcolor[HTML]{A0CBE2}0.77 & \cellcolor[HTML]{B7D4EA}0.52 & \cellcolor[HTML]{B7D4EA}0.51 & \cellcolor[HTML]{92C4DE}0.78 \\
     & Format & 60 & 333 & 3.56 & 6.15 & 37.77 & 21.44 & 59.21 & \cellcolor[HTML]{AEDEA7}134.44 & \cellcolor[HTML]{63BC6E}135.52 & \cellcolor[HTML]{ABD0E6}0.71 & \cellcolor[HTML]{ABD0E6}0.67 & \cellcolor[HTML]{B7D4EA}0.74 & \cellcolor[HTML]{ABD0E6}0.62 & \cellcolor[HTML]{ABD0E6}0.60 & \cellcolor[HTML]{B7D4EA}0.77 \\
     & Polite & 60 & 352 & 3.57 & 6.37 & 38.03 & 22.54 & 60.58 & \cellcolor[HTML]{B8E3B2}136.44 & \cellcolor[HTML]{7FC97F}143.81 & \cellcolor[HTML]{77B5D9}0.82 & \cellcolor[HTML]{69ADD5}0.78 & \cellcolor[HTML]{5DA5D1}0.85 & \cellcolor[HTML]{92C4DE}0.68 & \cellcolor[HTML]{77B5D9}0.69 & \cellcolor[HTML]{A0CBE2}0.78 \\
     & Technical & 59 & 360 & 3.56 & 6.11 & 37.14 & 23.36 & 60.50 & \cellcolor[HTML]{97D492}132.24 & \cellcolor[HTML]{8BCF89}147.93 & \cellcolor[HTML]{69ADD5}0.84 & \cellcolor[HTML]{5DA5D1}0.83 & \cellcolor[HTML]{92C4DE}0.79 & \cellcolor[HTML]{5DA5D1}0.74 & \cellcolor[HTML]{69ADD5}0.70 & \cellcolor[HTML]{77B5D9}0.82 \\
    \midrule
    \multirow{8}{*}{Qwen-2.5-0.5B} & Base & 50 & 234 & 3.85 & 5.22 & 12.83 & 15.04 & 27.87 & \cellcolor[HTML]{8BCF89}49.06 & \cellcolor[HTML]{63BC6E}80.25 & \cellcolor[HTML]{ABD0E6}0.69 & \cellcolor[HTML]{B7D4EA}0.60 & \cellcolor[HTML]{A0CBE2}0.73 & \cellcolor[HTML]{92C4DE}0.52 & \cellcolor[HTML]{77B5D9}0.55 & \cellcolor[HTML]{B7D4EA}0.73 \\
     & Persona & 50 & 253 & 3.81 & 5.13 & 12.51 & 16.48 & 28.99 & \cellcolor[HTML]{72C375}47.56 & \cellcolor[HTML]{97D492}88.09 & \cellcolor[HTML]{69ADD5}0.74 & \cellcolor[HTML]{5DA5D1}0.70 & \cellcolor[HTML]{5DA5D1}0.78 & \cellcolor[HTML]{77B5D9}0.57 & \cellcolor[HTML]{92C4DE}0.55 & \cellcolor[HTML]{77B5D9}0.82 \\
     & Aggressive & 50 & 275 & 3.78 & 5.05 & 12.57 & 17.99 & 30.55 & \cellcolor[HTML]{63BC6E}47.28 & \cellcolor[HTML]{AEDEA7}96.72 & \cellcolor[HTML]{5DA5D1}0.76 & \cellcolor[HTML]{77B5D9}0.65 & \cellcolor[HTML]{77B5D9}0.77 & \cellcolor[HTML]{5DA5D1}0.60 & \cellcolor[HTML]{69ADD5}0.57 & \cellcolor[HTML]{5DA5D1}0.87 \\
     & Conversational & 52 & 256 & 3.82 & 5.28 & 13.13 & 16.56 & 29.68 & \cellcolor[HTML]{97D492}49.89 & \cellcolor[HTML]{A3DA9D}89.12 & \cellcolor[HTML]{92C4DE}0.71 & \cellcolor[HTML]{ABD0E6}0.62 & \cellcolor[HTML]{B7D4EA}0.71 & \cellcolor[HTML]{ABD0E6}0.51 & \cellcolor[HTML]{B7D4EA}0.52 & \cellcolor[HTML]{85BCDC}0.81 \\
     & CoT & 53 & 243 & 3.75 & 5.03 & 13.61 & 15.77 & 29.39 & \cellcolor[HTML]{AEDEA7}50.94 & \cellcolor[HTML]{7FC97F}84.26 & \cellcolor[HTML]{A0CBE2}0.69 & \cellcolor[HTML]{ABD0E6}0.62 & \cellcolor[HTML]{85BCDC}0.74 & \cellcolor[HTML]{A0CBE2}0.52 & \cellcolor[HTML]{A0CBE2}0.53 & \cellcolor[HTML]{A0CBE2}0.77 \\
     & Format & 54 & 281 & 3.80 & 5.00 & 13.31 & 18.90 & 32.21 & \cellcolor[HTML]{A3DA9D}50.38 & \cellcolor[HTML]{B8E3B2}101.00 & \cellcolor[HTML]{77B5D9}0.72 & \cellcolor[HTML]{77B5D9}0.65 & \cellcolor[HTML]{B7D4EA}0.71 & \cellcolor[HTML]{69ADD5}0.58 & \cellcolor[HTML]{5DA5D1}0.61 & \cellcolor[HTML]{ABD0E6}0.77 \\
     & Polite & 54 & 253 & 3.82 & 5.27 & 13.46 & 16.29 & 29.76 & \cellcolor[HTML]{B8E3B2}51.12 & \cellcolor[HTML]{8BCF89}87.08 & \cellcolor[HTML]{85BCDC}0.71 & \cellcolor[HTML]{92C4DE}0.63 & \cellcolor[HTML]{92C4DE}0.73 & \cellcolor[HTML]{85BCDC}0.55 & \cellcolor[HTML]{ABD0E6}0.52 & \cellcolor[HTML]{92C4DE}0.79 \\
     & Technical & 53 & 233 & 3.79 & 4.94 & 12.94 & 15.28 & 28.22 & \cellcolor[HTML]{7FC97F}48.71 & \cellcolor[HTML]{72C375}81.31 & \cellcolor[HTML]{B7D4EA}0.68 & \cellcolor[HTML]{85BCDC}0.63 & \cellcolor[HTML]{77B5D9}0.77 & \cellcolor[HTML]{B7D4EA}0.51 & \cellcolor[HTML]{85BCDC}0.55 & \cellcolor[HTML]{69ADD5}0.87 \\
    \midrule
    \multirow{8}{*}{Qwen-2.5-1.5B} & Base & 50 & 272 & 3.54 & 5.12 & 43.47 & 30.70 & 74.17 & \cellcolor[HTML]{7FC97F}154.18 & \cellcolor[HTML]{97D492}155.54 & \cellcolor[HTML]{B7D4EA}0.71 & \cellcolor[HTML]{B7D4EA}0.68 & \cellcolor[HTML]{B7D4EA}0.76 & \cellcolor[HTML]{ABD0E6}0.61 & \cellcolor[HTML]{ABD0E6}0.61 & \cellcolor[HTML]{B7D4EA}0.78 \\
     & Persona & 50 & 294 & 3.55 & 5.08 & 42.15 & 34.37 & 76.53 & \cellcolor[HTML]{63BC6E}150.20 & \cellcolor[HTML]{B8E3B2}173.41 & \cellcolor[HTML]{5DA5D1}0.85 & \cellcolor[HTML]{69ADD5}0.79 & \cellcolor[HTML]{77B5D9}0.82 & \cellcolor[HTML]{77B5D9}0.69 & \cellcolor[HTML]{5DA5D1}0.70 & \cellcolor[HTML]{A0CBE2}0.84 \\
     & Aggressive & 50 & 259 & 3.55 & 5.00 & 42.92 & 29.50 & 72.42 & \cellcolor[HTML]{72C375}152.59 & \cellcolor[HTML]{8BCF89}148.93 & \cellcolor[HTML]{69ADD5}0.82 & \cellcolor[HTML]{5DA5D1}0.80 & \cellcolor[HTML]{69ADD5}0.84 & \cellcolor[HTML]{69ADD5}0.70 & \cellcolor[HTML]{77B5D9}0.69 & \cellcolor[HTML]{77B5D9}0.90 \\
     & Conversational & 52 & 287 & 3.56 & 5.07 & 43.94 & 32.93 & 76.87 & \cellcolor[HTML]{8BCF89}157.11 & \cellcolor[HTML]{AEDEA7}166.39 & \cellcolor[HTML]{A0CBE2}0.76 & \cellcolor[HTML]{77B5D9}0.78 & \cellcolor[HTML]{A0CBE2}0.80 & \cellcolor[HTML]{92C4DE}0.67 & \cellcolor[HTML]{92C4DE}0.67 & \cellcolor[HTML]{92C4DE}0.89 \\
     & CoT & 53 & 204 & 3.55 & 5.01 & 46.12 & 22.24 & 68.36 & \cellcolor[HTML]{A3DA9D}163.99 & \cellcolor[HTML]{72C375}113.42 & \cellcolor[HTML]{ABD0E6}0.72 & \cellcolor[HTML]{ABD0E6}0.71 & \cellcolor[HTML]{85BCDC}0.82 & \cellcolor[HTML]{B7D4EA}0.57 & \cellcolor[HTML]{B7D4EA}0.56 & \cellcolor[HTML]{92C4DE}0.89 \\
     & Format & 54 & 201 & 3.54 & 5.03 & 46.59 & 21.94 & 68.53 & \cellcolor[HTML]{AEDEA7}164.76 & \cellcolor[HTML]{63BC6E}111.73 & \cellcolor[HTML]{92C4DE}0.77 & \cellcolor[HTML]{85BCDC}0.78 & \cellcolor[HTML]{92C4DE}0.81 & \cellcolor[HTML]{A0CBE2}0.66 & \cellcolor[HTML]{A0CBE2}0.66 & \cellcolor[HTML]{5DA5D1}0.93 \\
     & Polite & 54 & 275 & 3.55 & 5.14 & 46.50 & 30.83 & 77.33 & \cellcolor[HTML]{B8E3B2}165.73 & \cellcolor[HTML]{A3DA9D}156.91 & \cellcolor[HTML]{77B5D9}0.79 & \cellcolor[HTML]{92C4DE}0.76 & \cellcolor[HTML]{ABD0E6}0.80 & \cellcolor[HTML]{5DA5D1}0.72 & \cellcolor[HTML]{69ADD5}0.70 & \cellcolor[HTML]{ABD0E6}0.83 \\
     & Technical & 53 & 259 & 3.57 & 5.12 & 44.02 & 28.73 & 72.75 & \cellcolor[HTML]{97D492}157.62 & \cellcolor[HTML]{7FC97F}146.32 & \cellcolor[HTML]{92C4DE}0.77 & \cellcolor[HTML]{A0CBE2}0.75 & \cellcolor[HTML]{5DA5D1}0.84 & \cellcolor[HTML]{85BCDC}0.69 & \cellcolor[HTML]{85BCDC}0.69 & \cellcolor[HTML]{69ADD5}0.93 \\
    \midrule
    \multirow{8}{*}{SmolLM2-360M} & Base & 62 & 243 & 3.82 & 4.80 & 7.92 & 13.55 & 21.48 & \cellcolor[HTML]{72C375}30.03 & \cellcolor[HTML]{7FC97F}67.24 & \cellcolor[HTML]{77B5D9}0.64 & \cellcolor[HTML]{85BCDC}0.55 & \cellcolor[HTML]{77B5D9}0.66 & \cellcolor[HTML]{77B5D9}0.49 & \cellcolor[HTML]{77B5D9}0.49 & \cellcolor[HTML]{92C4DE}0.68 \\
     & Persona & 63 & 201 & 3.81 & 4.67 & 7.98 & 11.16 & 19.15 & \cellcolor[HTML]{7FC97F}30.26 & \cellcolor[HTML]{63BC6E}54.40 & \cellcolor[HTML]{85BCDC}0.64 & \cellcolor[HTML]{69ADD5}0.56 & \cellcolor[HTML]{92C4DE}0.63 & \cellcolor[HTML]{69ADD5}0.50 & \cellcolor[HTML]{5DA5D1}0.54 & \cellcolor[HTML]{69ADD5}0.77 \\
     & Aggressive & 62 & 258 & 3.83 & 4.77 & 7.87 & 14.50 & 22.37 & \cellcolor[HTML]{63BC6E}29.96 & \cellcolor[HTML]{97D492}70.90 & \cellcolor[HTML]{92C4DE}0.62 & \cellcolor[HTML]{A0CBE2}0.49 & \cellcolor[HTML]{ABD0E6}0.60 & \cellcolor[HTML]{A0CBE2}0.38 & \cellcolor[HTML]{ABD0E6}0.44 & \cellcolor[HTML]{ABD0E6}0.61 \\
     & Conversational & 63 & 262 & 3.79 & 4.82 & 8.28 & 14.66 & 22.94 & \cellcolor[HTML]{97D492}31.20 & \cellcolor[HTML]{AEDEA7}72.21 & \cellcolor[HTML]{ABD0E6}0.58 & \cellcolor[HTML]{ABD0E6}0.44 & \cellcolor[HTML]{B7D4EA}0.58 & \cellcolor[HTML]{ABD0E6}0.37 & \cellcolor[HTML]{B7D4EA}0.38 & \cellcolor[HTML]{B7D4EA}0.60 \\
     & CoT & 65 & 259 & 3.81 & 4.78 & 8.18 & 14.54 & 22.72 & \cellcolor[HTML]{8BCF89}31.07 & \cellcolor[HTML]{A3DA9D}71.86 & \cellcolor[HTML]{B7D4EA}0.54 & \cellcolor[HTML]{B7D4EA}0.44 & \cellcolor[HTML]{A0CBE2}0.61 & \cellcolor[HTML]{92C4DE}0.39 & \cellcolor[HTML]{A0CBE2}0.45 & \cellcolor[HTML]{A0CBE2}0.64 \\
     & Format & 66 & 255 & 3.82 & 4.77 & 8.30 & 14.27 & 22.57 & \cellcolor[HTML]{B8E3B2}31.63 & \cellcolor[HTML]{8BCF89}70.13 & \cellcolor[HTML]{5DA5D1}0.72 & \cellcolor[HTML]{77B5D9}0.56 & \cellcolor[HTML]{69ADD5}0.66 & \cellcolor[HTML]{5DA5D1}0.54 & \cellcolor[HTML]{69ADD5}0.51 & \cellcolor[HTML]{77B5D9}0.74 \\
     & Polite & 66 & 218 & 3.81 & 4.84 & 8.32 & 12.09 & 20.41 & \cellcolor[HTML]{A3DA9D}31.42 & \cellcolor[HTML]{72C375}59.81 & \cellcolor[HTML]{69ADD5}0.65 & \cellcolor[HTML]{5DA5D1}0.62 & \cellcolor[HTML]{5DA5D1}0.68 & \cellcolor[HTML]{85BCDC}0.45 & \cellcolor[HTML]{85BCDC}0.47 & \cellcolor[HTML]{5DA5D1}0.80 \\
     & Technical & 64 & 269 & 3.81 & 4.76 & 8.28 & 15.15 & 23.44 & \cellcolor[HTML]{AEDEA7}31.45 & \cellcolor[HTML]{B8E3B2}75.47 & \cellcolor[HTML]{A0CBE2}0.61 & \cellcolor[HTML]{92C4DE}0.50 & \cellcolor[HTML]{85BCDC}0.65 & \cellcolor[HTML]{B7D4EA}0.36 & \cellcolor[HTML]{92C4DE}0.46 & \cellcolor[HTML]{85BCDC}0.69 \\
    \bottomrule
  \end{tabular}%
  }
\end{minipage}%
}
\end{table*}

%% file: tables/phrasing_pattern_overview_7.tex

\begin{table*}[p]
\centering
\rotatebox{90}{%
\begin{minipage}{0.92\textheight}
\centering
\captionof{table}{
Phrasing pattern summary across five LLMs on Pixel 7. Energy columns are shaded green, where darker indicates lower energy, and quality columns are shaded blue, where darker indicates higher quality.
}

\label{tab:phrasing-pattern-7-overview}

\scriptsize
\setlength{\tabcolsep}{2.2pt}
\renewcommand{\arraystretch}{0.9}

\resizebox{\linewidth}{!}{%
\begin{tabular}{llrrrrrrrrrrrrrrr}
\toprule
Model & Phrasing &
\makecell{Prompt\\Tok.} &
\makecell{Completion\\Tok.} &
\makecell{Prefill\\Power (W)} &
\makecell{Decode\\Power (W)} &
\makecell{Prefill\\Latency (s)} &
\makecell{Decode\\Latency (s)} &
\makecell{Total\\Latency (s)} &
\makecell{Prefill\\Energy (J)} &
\makecell{Decode\\Energy (J)} &
Rel. & Corr. & Coh. & Comp. &
\makecell{Instr.\\Adh.} &
\makecell{Internal\\Cons.} \\
\midrule
    \multirow{8}{*}{Gemma-2-2B} & Base & 41 & 538 & 2.79 & 3.13 & 14.95 & 107.76 & 122.71 & \cellcolor[HTML]{97D492}43.87 & \cellcolor[HTML]{AEDEA7}337.80 & \cellcolor[HTML]{A0CBE2}0.89 & \cellcolor[HTML]{77B5D9}0.90 & \cellcolor[HTML]{69ADD5}0.89 & \cellcolor[HTML]{69ADD5}0.84 & \cellcolor[HTML]{69ADD5}0.81 & \cellcolor[HTML]{77B5D9}0.95 \\
     & Persona & 41 & 533 & 2.84 & 3.15 & 14.34 & 107.88 & 122.22 & \cellcolor[HTML]{72C375}43.03 & \cellcolor[HTML]{A3DA9D}335.80 & \cellcolor[HTML]{B7D4EA}0.88 & \cellcolor[HTML]{92C4DE}0.88 & \cellcolor[HTML]{B7D4EA}0.85 & \cellcolor[HTML]{B7D4EA}0.79 & \cellcolor[HTML]{ABD0E6}0.77 & \cellcolor[HTML]{ABD0E6}0.92 \\
     & Aggressive & 40 & 556 & 2.75 & 3.13 & 13.58 & 112.61 & 126.19 & \cellcolor[HTML]{63BC6E}38.96 & \cellcolor[HTML]{B8E3B2}350.58 & \cellcolor[HTML]{ABD0E6}0.89 & \cellcolor[HTML]{B7D4EA}0.86 & \cellcolor[HTML]{ABD0E6}0.86 & \cellcolor[HTML]{92C4DE}0.81 & \cellcolor[HTML]{85BCDC}0.80 & \cellcolor[HTML]{92C4DE}0.94 \\
     & Conversational & 42 & 519 & 2.84 & 3.15 & 14.64 & 103.74 & 118.38 & \cellcolor[HTML]{7FC97F}43.31 & \cellcolor[HTML]{97D492}324.21 & \cellcolor[HTML]{5DA5D1}0.92 & \cellcolor[HTML]{69ADD5}0.92 & \cellcolor[HTML]{77B5D9}0.89 & \cellcolor[HTML]{85BCDC}0.82 & \cellcolor[HTML]{77B5D9}0.81 & \cellcolor[HTML]{5DA5D1}0.98 \\
     & CoT & 44 & 518 & 2.91 & 3.13 & 15.60 & 103.27 & 118.87 & \cellcolor[HTML]{AEDEA7}46.89 & \cellcolor[HTML]{8BCF89}323.56 & \cellcolor[HTML]{A0CBE2}0.89 & \cellcolor[HTML]{85BCDC}0.88 & \cellcolor[HTML]{ABD0E6}0.86 & \cellcolor[HTML]{A0CBE2}0.80 & \cellcolor[HTML]{B7D4EA}0.76 & \cellcolor[HTML]{A0CBE2}0.93 \\
     & Format & 45 & 476 & 2.94 & 3.11 & 15.66 & 94.34 & 110.00 & \cellcolor[HTML]{B8E3B2}47.28 & \cellcolor[HTML]{72C375}294.89 & \cellcolor[HTML]{69ADD5}0.92 & \cellcolor[HTML]{5DA5D1}0.93 & \cellcolor[HTML]{5DA5D1}0.91 & \cellcolor[HTML]{5DA5D1}0.86 & \cellcolor[HTML]{92C4DE}0.79 & \cellcolor[HTML]{69ADD5}0.96 \\
     & Polite & 44 & 485 & 2.92 & 3.17 & 14.88 & 95.43 & 110.31 & \cellcolor[HTML]{A3DA9D}44.91 & \cellcolor[HTML]{7FC97F}301.51 & \cellcolor[HTML]{85BCDC}0.91 & \cellcolor[HTML]{ABD0E6}0.87 & \cellcolor[HTML]{85BCDC}0.88 & \cellcolor[HTML]{B7D4EA}0.79 & \cellcolor[HTML]{5DA5D1}0.83 & \cellcolor[HTML]{85BCDC}0.95 \\
     & Technical & 43 & 473 & 2.84 & 3.08 & 14.74 & 94.32 & 109.06 & \cellcolor[HTML]{8BCF89}43.48 & \cellcolor[HTML]{63BC6E}294.37 & \cellcolor[HTML]{85BCDC}0.91 & \cellcolor[HTML]{A0CBE2}0.87 & \cellcolor[HTML]{92C4DE}0.87 & \cellcolor[HTML]{77B5D9}0.83 & \cellcolor[HTML]{ABD0E6}0.77 & \cellcolor[HTML]{B7D4EA}0.91 \\
    \midrule
    \multirow{8}{*}{LLaMA-3.2-1B} & Base & 56 & 381 & 2.47 & 3.39 & 55.09 & 35.88 & 90.97 & \cellcolor[HTML]{7FC97F}138.13 & \cellcolor[HTML]{97D492}122.21 & \cellcolor[HTML]{77B5D9}0.81 & \cellcolor[HTML]{ABD0E6}0.68 & \cellcolor[HTML]{A0CBE2}0.80 & \cellcolor[HTML]{69ADD5}0.72 & \cellcolor[HTML]{92C4DE}0.66 & \cellcolor[HTML]{A0CBE2}0.80 \\
     & Persona & 56 & 410 & 2.48 & 3.40 & 54.50 & 38.94 & 93.44 & \cellcolor[HTML]{63BC6E}136.55 & \cellcolor[HTML]{B8E3B2}133.09 & \cellcolor[HTML]{69ADD5}0.83 & \cellcolor[HTML]{77B5D9}0.77 & \cellcolor[HTML]{77B5D9}0.81 & \cellcolor[HTML]{77B5D9}0.71 & \cellcolor[HTML]{5DA5D1}0.73 & \cellcolor[HTML]{5DA5D1}0.89 \\
     & Aggressive & 56 & 355 & 2.49 & 3.40 & 54.89 & 33.48 & 88.37 & \cellcolor[HTML]{72C375}137.36 & \cellcolor[HTML]{7FC97F}114.63 & \cellcolor[HTML]{85BCDC}0.80 & \cellcolor[HTML]{85BCDC}0.71 & \cellcolor[HTML]{A0CBE2}0.80 & \cellcolor[HTML]{ABD0E6}0.64 & \cellcolor[HTML]{A0CBE2}0.62 & \cellcolor[HTML]{85BCDC}0.82 \\
     & Conversational & 58 & 387 & 2.48 & 3.40 & 57.38 & 36.52 & 93.90 & \cellcolor[HTML]{8BCF89}144.02 & \cellcolor[HTML]{AEDEA7}125.17 & \cellcolor[HTML]{A0CBE2}0.78 & \cellcolor[HTML]{ABD0E6}0.68 & \cellcolor[HTML]{85BCDC}0.81 & \cellcolor[HTML]{92C4DE}0.69 & \cellcolor[HTML]{85BCDC}0.66 & \cellcolor[HTML]{ABD0E6}0.79 \\
     & CoT & 59 & 376 & 2.49 & 3.36 & 58.56 & 35.74 & 94.30 & \cellcolor[HTML]{97D492}147.47 & \cellcolor[HTML]{A3DA9D}122.24 & \cellcolor[HTML]{B7D4EA}0.63 & \cellcolor[HTML]{B7D4EA}0.64 & \cellcolor[HTML]{ABD0E6}0.77 & \cellcolor[HTML]{B7D4EA}0.54 & \cellcolor[HTML]{B7D4EA}0.56 & \cellcolor[HTML]{B7D4EA}0.79 \\
     & Format & 60 & 334 & 2.49 & 3.33 & 60.16 & 31.61 & 91.77 & \cellcolor[HTML]{AEDEA7}151.09 & \cellcolor[HTML]{63BC6E}107.99 & \cellcolor[HTML]{ABD0E6}0.72 & \cellcolor[HTML]{92C4DE}0.68 & \cellcolor[HTML]{B7D4EA}0.74 & \cellcolor[HTML]{A0CBE2}0.66 & \cellcolor[HTML]{ABD0E6}0.60 & \cellcolor[HTML]{92C4DE}0.80 \\
     & Polite & 60 & 350 & 2.50 & 3.43 & 60.44 & 32.85 & 93.29 & \cellcolor[HTML]{B8E3B2}152.19 & \cellcolor[HTML]{72C375}112.44 & \cellcolor[HTML]{92C4DE}0.79 & \cellcolor[HTML]{77B5D9}0.77 & \cellcolor[HTML]{5DA5D1}0.84 & \cellcolor[HTML]{92C4DE}0.69 & \cellcolor[HTML]{77B5D9}0.68 & \cellcolor[HTML]{77B5D9}0.83 \\
     & Technical & 59 & 360 & 2.48 & 3.29 & 58.93 & 34.11 & 93.04 & \cellcolor[HTML]{A3DA9D}147.78 & \cellcolor[HTML]{8BCF89}116.49 & \cellcolor[HTML]{5DA5D1}0.84 & \cellcolor[HTML]{5DA5D1}0.86 & \cellcolor[HTML]{69ADD5}0.82 & \cellcolor[HTML]{5DA5D1}0.74 & \cellcolor[HTML]{69ADD5}0.69 & \cellcolor[HTML]{69ADD5}0.86 \\
    \midrule
    \multirow{8}{*}{Qwen-2.5-0.5B} & Base & 50 & 234 & 2.12 & 2.69 & 17.70 & 27.13 & 44.83 & \cellcolor[HTML]{7FC97F}37.99 & \cellcolor[HTML]{63BC6E}72.57 & \cellcolor[HTML]{ABD0E6}0.69 & \cellcolor[HTML]{B7D4EA}0.61 & \cellcolor[HTML]{B7D4EA}0.72 & \cellcolor[HTML]{A0CBE2}0.55 & \cellcolor[HTML]{92C4DE}0.55 & \cellcolor[HTML]{B7D4EA}0.74 \\
     & Persona & 50 & 253 & 2.13 & 2.66 & 17.16 & 29.66 & 46.82 & \cellcolor[HTML]{63BC6E}36.78 & \cellcolor[HTML]{97D492}78.62 & \cellcolor[HTML]{69ADD5}0.74 & \cellcolor[HTML]{5DA5D1}0.71 & \cellcolor[HTML]{5DA5D1}0.77 & \cellcolor[HTML]{69ADD5}0.56 & \cellcolor[HTML]{69ADD5}0.58 & \cellcolor[HTML]{77B5D9}0.85 \\
     & Aggressive & 50 & 275 & 2.13 & 2.66 & 17.18 & 32.19 & 49.38 & \cellcolor[HTML]{72C375}36.83 & \cellcolor[HTML]{AEDEA7}85.82 & \cellcolor[HTML]{5DA5D1}0.75 & \cellcolor[HTML]{85BCDC}0.66 & \cellcolor[HTML]{77B5D9}0.75 & \cellcolor[HTML]{5DA5D1}0.61 & \cellcolor[HTML]{5DA5D1}0.58 & \cellcolor[HTML]{5DA5D1}0.87 \\
     & Conversational & 52 & 256 & 2.12 & 2.68 & 18.06 & 29.70 & 47.76 & \cellcolor[HTML]{97D492}38.62 & \cellcolor[HTML]{A3DA9D}79.46 & \cellcolor[HTML]{ABD0E6}0.69 & \cellcolor[HTML]{A0CBE2}0.64 & \cellcolor[HTML]{A0CBE2}0.73 & \cellcolor[HTML]{ABD0E6}0.50 & \cellcolor[HTML]{B7D4EA}0.53 & \cellcolor[HTML]{85BCDC}0.81 \\
     & CoT & 53 & 243 & 2.14 & 2.65 & 18.88 & 28.44 & 47.32 & \cellcolor[HTML]{B8E3B2}40.71 & \cellcolor[HTML]{7FC97F}75.84 & \cellcolor[HTML]{85BCDC}0.72 & \cellcolor[HTML]{69ADD5}0.68 & \cellcolor[HTML]{85BCDC}0.74 & \cellcolor[HTML]{85BCDC}0.55 & \cellcolor[HTML]{92C4DE}0.55 & \cellcolor[HTML]{92C4DE}0.77 \\
     & Format & 54 & 281 & 2.15 & 2.64 & 18.42 & 34.05 & 52.47 & \cellcolor[HTML]{A3DA9D}39.97 & \cellcolor[HTML]{B8E3B2}90.27 & \cellcolor[HTML]{77B5D9}0.73 & \cellcolor[HTML]{77B5D9}0.67 & \cellcolor[HTML]{A0CBE2}0.73 & \cellcolor[HTML]{77B5D9}0.56 & \cellcolor[HTML]{77B5D9}0.56 & \cellcolor[HTML]{A0CBE2}0.77 \\
     & Polite & 54 & 253 & 2.14 & 2.68 & 18.69 & 29.21 & 47.90 & \cellcolor[HTML]{AEDEA7}40.42 & \cellcolor[HTML]{8BCF89}78.12 & \cellcolor[HTML]{92C4DE}0.71 & \cellcolor[HTML]{92C4DE}0.65 & \cellcolor[HTML]{ABD0E6}0.73 & \cellcolor[HTML]{A0CBE2}0.55 & \cellcolor[HTML]{ABD0E6}0.54 & \cellcolor[HTML]{ABD0E6}0.75 \\
     & Technical & 53 & 233 & 2.12 & 2.61 & 17.78 & 27.52 & 45.30 & \cellcolor[HTML]{8BCF89}38.13 & \cellcolor[HTML]{72C375}73.23 & \cellcolor[HTML]{B7D4EA}0.65 & \cellcolor[HTML]{ABD0E6}0.62 & \cellcolor[HTML]{69ADD5}0.77 & \cellcolor[HTML]{B7D4EA}0.48 & \cellcolor[HTML]{A0CBE2}0.55 & \cellcolor[HTML]{69ADD5}0.86 \\
    \midrule
    \multirow{8}{*}{Qwen-2.5-1.5B} & Base & 50 & 272 & 2.33 & 2.86 & 65.33 & 44.36 & 109.69 & \cellcolor[HTML]{7FC97F}154.26 & \cellcolor[HTML]{97D492}125.04 & \cellcolor[HTML]{92C4DE}0.77 & \cellcolor[HTML]{B7D4EA}0.67 & \cellcolor[HTML]{B7D4EA}0.77 & \cellcolor[HTML]{ABD0E6}0.62 & \cellcolor[HTML]{ABD0E6}0.64 & \cellcolor[HTML]{B7D4EA}0.76 \\
     & Persona & 50 & 294 & 2.34 & 2.83 & 62.71 & 49.72 & 112.44 & \cellcolor[HTML]{63BC6E}148.01 & \cellcolor[HTML]{B8E3B2}138.44 & \cellcolor[HTML]{69ADD5}0.81 & \cellcolor[HTML]{7CB7DA}0.75 & \cellcolor[HTML]{ABD0E6}0.78 & \cellcolor[HTML]{85BCDC}0.69 & \cellcolor[HTML]{77B5D9}0.69 & \cellcolor[HTML]{ABD0E6}0.84 \\
     & Aggressive & 50 & 259 & 2.33 & 2.82 & 63.87 & 42.68 & 106.55 & \cellcolor[HTML]{72C375}150.01 & \cellcolor[HTML]{8BCF89}119.82 & \cellcolor[HTML]{5DA5D1}0.84 & \cellcolor[HTML]{5DA5D1}0.79 & \cellcolor[HTML]{77B5D9}0.84 & \cellcolor[HTML]{5DA5D1}0.71 & \cellcolor[HTML]{5DA5D1}0.71 & \cellcolor[HTML]{77B5D9}0.90 \\
     & Conversational & 52 & 287 & 2.35 & 2.87 & 66.03 & 47.59 & 113.62 & \cellcolor[HTML]{8BCF89}156.63 & \cellcolor[HTML]{AEDEA7}134.89 & \cellcolor[HTML]{A0CBE2}0.77 & \cellcolor[HTML]{6AAED6}0.78 & \cellcolor[HTML]{85BCDC}0.81 & \cellcolor[HTML]{92C4DE}0.69 & \cellcolor[HTML]{A0CBE2}0.65 & \cellcolor[HTML]{92C4DE}0.86 \\
     & CoT & 53 & 204 & 2.35 & 2.81 & 70.27 & 32.11 & 102.39 & \cellcolor[HTML]{AEDEA7}166.26 & \cellcolor[HTML]{72C375}90.27 & \cellcolor[HTML]{B7D4EA}0.73 & \cellcolor[HTML]{A9CFE5}0.70 & \cellcolor[HTML]{A0CBE2}0.81 & \cellcolor[HTML]{B7D4EA}0.61 & \cellcolor[HTML]{B7D4EA}0.54 & \cellcolor[HTML]{85BCDC}0.88 \\
     & Format & 54 & 201 & 2.34 & 2.83 & 70.20 & 31.73 & 101.93 & \cellcolor[HTML]{A3DA9D}165.84 & \cellcolor[HTML]{63BC6E}89.56 & \cellcolor[HTML]{ABD0E6}0.76 & \cellcolor[HTML]{5DA5D1}0.79 & \cellcolor[HTML]{5DA5D1}0.85 & \cellcolor[HTML]{A0CBE2}0.64 & \cellcolor[HTML]{85BCDC}0.69 & \cellcolor[HTML]{69ADD5}0.92 \\
     & Polite & 54 & 275 & 2.35 & 2.85 & 70.63 & 44.64 & 115.26 & \cellcolor[HTML]{B8E3B2}166.84 & \cellcolor[HTML]{A3DA9D}125.90 & \cellcolor[HTML]{77B5D9}0.81 & \cellcolor[HTML]{8CC0DD}0.74 & \cellcolor[HTML]{A0CBE2}0.81 & \cellcolor[HTML]{69ADD5}0.71 & \cellcolor[HTML]{92C4DE}0.68 & \cellcolor[HTML]{A0CBE2}0.85 \\
     & Technical & 53 & 259 & 2.34 & 2.87 & 66.27 & 41.53 & 107.80 & \cellcolor[HTML]{97D492}156.72 & \cellcolor[HTML]{7FC97F}118.12 & \cellcolor[HTML]{85BCDC}0.80 & \cellcolor[HTML]{9CC9E1}0.74 & \cellcolor[HTML]{69ADD5}0.85 & \cellcolor[HTML]{77B5D9}0.70 & \cellcolor[HTML]{69ADD5}0.70 & \cellcolor[HTML]{5DA5D1}0.93 \\
    \midrule
    \multirow{8}{*}{SmolLM2-360M} & Base & 62 & 243 & 2.05 & 2.45 & 11.07 & 16.89 & 27.96 & \cellcolor[HTML]{7FC97F}22.85 & \cellcolor[HTML]{7FC97F}41.36 & \cellcolor[HTML]{92C4DE}0.64 & \cellcolor[HTML]{77B5D9}0.58 & \cellcolor[HTML]{92C4DE}0.63 & \cellcolor[HTML]{69ADD5}0.48 & \cellcolor[HTML]{5DA5D1}0.51 & \cellcolor[HTML]{85BCDC}0.69 \\
     & Persona & 63 & 201 & 2.04 & 2.43 & 11.06 & 13.65 & 24.71 & \cellcolor[HTML]{72C375}22.73 & \cellcolor[HTML]{63BC6E}33.03 & \cellcolor[HTML]{69ADD5}0.64 & \cellcolor[HTML]{69ADD5}0.60 & \cellcolor[HTML]{85BCDC}0.64 & \cellcolor[HTML]{77B5D9}0.46 & \cellcolor[HTML]{69ADD5}0.49 & \cellcolor[HTML]{5DA5D1}0.73 \\
     & Aggressive & 62 & 271 & 2.07 & 2.47 & 10.93 & 19.58 & 30.51 & \cellcolor[HTML]{63BC6E}22.72 & \cellcolor[HTML]{AEDEA7}47.68 & \cellcolor[HTML]{A0CBE2}0.61 & \cellcolor[HTML]{B7D4EA}0.44 & \cellcolor[HTML]{ABD0E6}0.58 & \cellcolor[HTML]{92C4DE}0.39 & \cellcolor[HTML]{ABD0E6}0.42 & \cellcolor[HTML]{ABD0E6}0.61 \\
     & Conversational & 63 & 265 & 2.05 & 2.49 & 11.55 & 18.53 & 30.08 & \cellcolor[HTML]{A3DA9D}23.83 & \cellcolor[HTML]{97D492}45.88 & \cellcolor[HTML]{ABD0E6}0.55 & \cellcolor[HTML]{ABD0E6}0.45 & \cellcolor[HTML]{B7D4EA}0.54 & \cellcolor[HTML]{ABD0E6}0.38 & \cellcolor[HTML]{B7D4EA}0.36 & \cellcolor[HTML]{B7D4EA}0.58 \\
     & CoT & 65 & 268 & 2.04 & 2.47 & 11.36 & 19.13 & 30.50 & \cellcolor[HTML]{8BCF89}23.31 & \cellcolor[HTML]{A3DA9D}46.52 & \cellcolor[HTML]{B7D4EA}0.53 & \cellcolor[HTML]{A0CBE2}0.47 & \cellcolor[HTML]{A0CBE2}0.62 & \cellcolor[HTML]{B7D4EA}0.37 & \cellcolor[HTML]{A0CBE2}0.46 & \cellcolor[HTML]{A0CBE2}0.61 \\
     & Format & 66 & 255 & 2.05 & 2.46 & 11.59 & 17.85 & 29.44 & \cellcolor[HTML]{97D492}23.79 & \cellcolor[HTML]{8BCF89}43.87 & \cellcolor[HTML]{5DA5D1}0.74 & \cellcolor[HTML]{85BCDC}0.54 & \cellcolor[HTML]{5DA5D1}0.70 & \cellcolor[HTML]{5DA5D1}0.56 & \cellcolor[HTML]{A0CBE2}0.46 & \cellcolor[HTML]{92C4DE}0.68 \\
     & Polite & 66 & 218 & 2.09 & 2.44 & 11.63 & 14.80 & 26.43 & \cellcolor[HTML]{B8E3B2}24.20 & \cellcolor[HTML]{72C375}36.19 & \cellcolor[HTML]{92C4DE}0.64 & \cellcolor[HTML]{5DA5D1}0.63 & \cellcolor[HTML]{69ADD5}0.67 & \cellcolor[HTML]{85BCDC}0.45 & \cellcolor[HTML]{77B5D9}0.48 & \cellcolor[HTML]{69ADD5}0.73 \\
     & Technical & 64 & 278 & 2.07 & 2.47 & 11.55 & 20.10 & 31.65 & \cellcolor[HTML]{AEDEA7}23.97 & \cellcolor[HTML]{B8E3B2}49.03 & \cellcolor[HTML]{92C4DE}0.64 & \cellcolor[HTML]{92C4DE}0.51 & \cellcolor[HTML]{77B5D9}0.65 & \cellcolor[HTML]{ABD0E6}0.38 & \cellcolor[HTML]{85BCDC}0.46 & \cellcolor[HTML]{85BCDC}0.69 \\
    \bottomrule
  \end{tabular}%
  }
\end{minipage}%
}
\end{table*}

%% file: tables/cognitive_load_ai2_arc_overview_7.tex
\begin{table}[p]
  \centering
  \caption{Cognitive load summary on AI2-ARC across five LLMs on Pixel 8 Pro. Within each model block, the energy columns are shaded green (darker = lower / better) and the accuracy column is shaded blue (darker = higher / better).}
  \label{tab:cognitive-load-ai2-arc-overview}
  \tiny
  \setlength{\tabcolsep}{2.5pt}
  \resizebox{\textwidth}{!}{%
  \begin{tabular}{llrrrrrrrrrr}
    \toprule
    Model & Load & Prompt Tokens & Completion Tokens & Prefill Power (W) & Decode Power (W) & Prefill Latency (s) & Decode Latency (s) & Total Latency (s) & Prefill Energy (J) & Decode Energy (J) & Accuracy \\
    \midrule
    \multirow{4}{*}{Gemma-2-2B} & Base & 63 & 174 & 3.95 & 6.37 & 6.63 & 16.66 & 23.28 & \cellcolor[HTML]{63BC6E}26.27 & \cellcolor[HTML]{63BC6E}106.04 & \cellcolor[HTML]{9CC9E1}0.84 \\
     & Intrinsic & 129 & 258 & 4.49 & 6.19 & 8.90 & 26.58 & 35.49 & \cellcolor[HTML]{9FD899}40.05 & \cellcolor[HTML]{B8E3B2}164.25 & \cellcolor[HTML]{9CC9E1}0.84 \\
     & Extraneous & 167 & 254 & 4.70 & 6.18 & 10.41 & 26.48 & 36.89 & \cellcolor[HTML]{B8E3B2}48.98 & \cellcolor[HTML]{9FD899}163.37 & \cellcolor[HTML]{B7D4EA}0.64 \\
     & Germane & 70 & 218 & 4.01 & 6.33 & 6.81 & 21.18 & 28.00 & \cellcolor[HTML]{83CB82}27.45 & \cellcolor[HTML]{83CB82}133.97 & \cellcolor[HTML]{5DA5D1}0.88 \\
    \midrule
    \multirow{4}{*}{LLaMA-3.2-1B} & Base & 79 & 197 & 3.59 & 6.04 & 47.06 & 11.95 & 59.00 & \cellcolor[HTML]{63BC6E}169.74 & \cellcolor[HTML]{63BC6E}72.67 & \cellcolor[HTML]{5DA5D1}0.64 \\
     & Intrinsic & 143 & 305 & 3.69 & 6.09 & 81.33 & 19.13 & 100.46 & \cellcolor[HTML]{9FD899}300.84 & \cellcolor[HTML]{B8E3B2}117.01 & \cellcolor[HTML]{B7D4EA}0.47 \\
     & Extraneous & 179 & 240 & 3.71 & 5.92 & 100.11 & 15.09 & 115.21 & \cellcolor[HTML]{B8E3B2}372.55 & \cellcolor[HTML]{83CB82}90.85 & \cellcolor[HTML]{9CC9E1}0.56 \\
     & Germane & 86 & 265 & 3.61 & 6.12 & 50.44 & 16.24 & 66.67 & \cellcolor[HTML]{83CB82}182.63 & \cellcolor[HTML]{9FD899}99.56 & \cellcolor[HTML]{7CB7DA}0.59 \\
    \midrule
    \multirow{4}{*}{Qwen-2.5-0.5B} & Base & 73 & 190 & 3.34 & 5.20 & 18.26 & 12.03 & 30.29 & \cellcolor[HTML]{63BC6E}60.95 & \cellcolor[HTML]{63BC6E}65.41 & \cellcolor[HTML]{9CC9E1}0.48 \\
     & Intrinsic & 138 & 207 & 3.40 & 5.27 & 31.38 & 13.60 & 44.98 & \cellcolor[HTML]{9FD899}106.70 & \cellcolor[HTML]{83CB82}73.30 & \cellcolor[HTML]{B7D4EA}0.44 \\
     & Extraneous & 179 & 258 & 3.41 & 5.33 & 39.15 & 17.30 & 56.44 & \cellcolor[HTML]{B8E3B2}133.53 & \cellcolor[HTML]{B8E3B2}92.93 & \cellcolor[HTML]{9CC9E1}0.48 \\
     & Germane & 80 & 238 & 3.35 & 5.34 & 19.71 & 15.44 & 35.15 & \cellcolor[HTML]{83CB82}66.10 & \cellcolor[HTML]{9FD899}84.44 & \cellcolor[HTML]{5DA5D1}0.52 \\
    \midrule
    \multirow{4}{*}{Qwen-2.5-1.5B} & Base & 73 & 125 & 3.59 & 5.65 & 58.55 & 12.23 & 70.79 & \cellcolor[HTML]{63BC6E}210.88 & \cellcolor[HTML]{63BC6E}71.92 & \cellcolor[HTML]{5DA5D1}0.80 \\
     & Intrinsic & 138 & 253 & 3.67 & 5.69 & 107.14 & 27.03 & 134.18 & \cellcolor[HTML]{9FD899}393.95 & \cellcolor[HTML]{9FD899}155.01 & \cellcolor[HTML]{B7D4EA}0.72 \\
     & Extraneous & 179 & 272 & 3.73 & 5.65 & 131.42 & 30.65 & 162.07 & \cellcolor[HTML]{B8E3B2}491.30 & \cellcolor[HTML]{B8E3B2}173.18 & \cellcolor[HTML]{7CB7DA}0.77 \\
     & Germane & 80 & 191 & 3.61 & 5.94 & 64.02 & 18.90 & 82.92 & \cellcolor[HTML]{83CB82}231.55 & \cellcolor[HTML]{83CB82}111.60 & \cellcolor[HTML]{9CC9E1}0.76 \\
    \midrule
    \multirow{4}{*}{SmolLM2-360M} & Base & 85 & 72 & 3.49 & 4.01 & 11.89 & 4.60 & 16.49 & \cellcolor[HTML]{63BC6E}41.63 & \cellcolor[HTML]{63BC6E}21.61 & \cellcolor[HTML]{5DA5D1}0.24 \\
     & Intrinsic & 150 & 160 & 3.46 & 4.63 & 21.09 & 10.31 & 31.40 & \cellcolor[HTML]{9FD899}73.13 & \cellcolor[HTML]{B8E3B2}49.85 & \cellcolor[HTML]{B7D4EA}0.17 \\
     & Extraneous & 190 & 122 & 3.47 & 4.46 & 26.94 & 7.89 & 34.83 & \cellcolor[HTML]{B8E3B2}93.58 & \cellcolor[HTML]{9FD899}37.92 & \cellcolor[HTML]{5DA5D1}0.24 \\
     & Germane & 92 & 115 & 3.49 & 4.45 & 12.85 & 7.40 & 20.25 & \cellcolor[HTML]{83CB82}44.83 & \cellcolor[HTML]{83CB82}34.94 & \cellcolor[HTML]{8CC0DD}0.20 \\
    \bottomrule
  \end{tabular}%
  }
\end{table}

%% file: tables/cognitive_load_boolq_overview_7.tex
\begin{table}[p]
  \centering
  \caption{Cognitive load summary on BoolQ across five LLMs on Pixel 8 Pro. Within each model block, the energy columns are shaded green (darker = lower / better) and the accuracy column is shaded blue (darker = higher / better).}
  \label{tab:cognitive-load-boolq-overview}
  \tiny
  \setlength{\tabcolsep}{2.5pt}
  \resizebox{\textwidth}{!}{%
  \begin{tabular}{llrrrrrrrrrr}
    \toprule
    Model & Load & Prompt Tokens & Completion Tokens & Prefill Power (W) & Decode Power (W) & Prefill Latency (s) & Decode Latency (s) & Total Latency (s) & Prefill Energy (J) & Decode Energy (J) & Accuracy \\
    \midrule
    \multirow{4}{*}{Gemma-2-2B} & Base & 51 & 119 & 4.15 & 6.15 & 6.23 & 11.56 & 17.79 & \cellcolor[HTML]{83CB82}25.86 & \cellcolor[HTML]{63BC6E}72.90 & \cellcolor[HTML]{5DA5D1}0.87 \\
     & Intrinsic & 100 & 137 & 4.57 & 6.11 & 7.97 & 13.86 & 21.83 & \cellcolor[HTML]{9FD899}36.45 & \cellcolor[HTML]{83CB82}86.23 & \cellcolor[HTML]{7CB7DA}0.72 \\
     & Extraneous & 132 & 236 & 4.86 & 6.21 & 9.14 & 24.91 & 34.05 & \cellcolor[HTML]{B8E3B2}44.54 & \cellcolor[HTML]{B8E3B2}154.16 & \cellcolor[HTML]{B7D4EA}0.57 \\
     & Germane & 48 & 215 & 4.06 & 6.37 & 6.10 & 21.49 & 27.59 & \cellcolor[HTML]{63BC6E}24.80 & \cellcolor[HTML]{9FD899}136.30 & \cellcolor[HTML]{9CC9E1}0.69 \\
    \midrule
    \multirow{4}{*}{LLaMA-3.2-1B} & Base & 65 & 116 & 3.68 & 5.94 & 40.47 & 7.10 & 47.57 & \cellcolor[HTML]{83CB82}149.01 & \cellcolor[HTML]{63BC6E}44.48 & \cellcolor[HTML]{7CB7DA}0.40 \\
     & Intrinsic & 113 & 145 & 3.78 & 6.14 & 67.84 & 9.10 & 76.93 & \cellcolor[HTML]{9FD899}256.49 & \cellcolor[HTML]{83CB82}56.83 & \cellcolor[HTML]{5DA5D1}0.49 \\
     & Extraneous & 141 & 236 & 3.81 & 6.17 & 82.20 & 15.12 & 97.31 & \cellcolor[HTML]{B8E3B2}312.80 & \cellcolor[HTML]{B8E3B2}94.92 & \cellcolor[HTML]{B7D4EA}0.32 \\
     & Germane & 63 & 220 & 3.69 & 6.36 & 39.75 & 13.77 & 53.52 & \cellcolor[HTML]{63BC6E}146.77 & \cellcolor[HTML]{9FD899}88.42 & \cellcolor[HTML]{9CC9E1}0.37 \\
    \midrule
    \multirow{4}{*}{Qwen-2.5-0.5B} & Base & 61 & 66 & 3.54 & 4.54 & 15.94 & 3.98 & 19.92 & \cellcolor[HTML]{83CB82}56.39 & \cellcolor[HTML]{63BC6E}19.20 & \cellcolor[HTML]{7CB7DA}0.60 \\
     & Intrinsic & 108 & 78 & 3.52 & 4.19 & 25.56 & 4.92 & 30.48 & \cellcolor[HTML]{9FD899}89.92 & \cellcolor[HTML]{83CB82}25.07 & \cellcolor[HTML]{5DA5D1}0.65 \\
     & Extraneous & 142 & 189 & 3.56 & 5.35 & 33.02 & 12.42 & 45.44 & \cellcolor[HTML]{B8E3B2}117.64 & \cellcolor[HTML]{B8E3B2}68.56 & \cellcolor[HTML]{B7D4EA}0.44 \\
     & Germane & 58 & 190 & 3.56 & 5.16 & 14.93 & 12.10 & 27.03 & \cellcolor[HTML]{63BC6E}52.96 & \cellcolor[HTML]{9FD899}67.07 & \cellcolor[HTML]{B7D4EA}0.44 \\
    \midrule
    \multirow{4}{*}{Qwen-2.5-1.5B} & Base & 61 & 56 & 3.59 & 4.90 & 47.75 & 4.84 & 52.59 & \cellcolor[HTML]{83CB82}171.70 & \cellcolor[HTML]{63BC6E}25.56 & \cellcolor[HTML]{5DA5D1}0.88 \\
     & Intrinsic & 108 & 101 & 3.64 & 5.11 & 81.40 & 9.38 & 90.78 & \cellcolor[HTML]{9FD899}296.11 & \cellcolor[HTML]{83CB82}52.01 & \cellcolor[HTML]{9CC9E1}0.67 \\
     & Extraneous & 142 & 166 & 3.67 & 5.46 & 106.24 & 16.81 & 123.06 & \cellcolor[HTML]{B8E3B2}389.55 & \cellcolor[HTML]{9FD899}93.43 & \cellcolor[HTML]{B7D4EA}0.65 \\
     & Germane & 58 & 192 & 3.63 & 5.69 & 45.98 & 18.08 & 64.06 & \cellcolor[HTML]{63BC6E}166.74 & \cellcolor[HTML]{B8E3B2}105.24 & \cellcolor[HTML]{7CB7DA}0.79 \\
    \midrule
    \multirow{4}{*}{SmolLM2-360M} & Base & 74 & 45 & 3.65 & 4.11 & 9.60 & 2.59 & 12.19 & \cellcolor[HTML]{83CB82}35.03 & \cellcolor[HTML]{63BC6E}12.30 & \cellcolor[HTML]{7CB7DA}0.49 \\
     & Intrinsic & 122 & 104 & 3.62 & 4.52 & 15.88 & 6.36 & 22.23 & \cellcolor[HTML]{9FD899}57.37 & \cellcolor[HTML]{9FD899}30.75 & \cellcolor[HTML]{5DA5D1}0.53 \\
     & Extraneous & 155 & 122 & 3.58 & 4.68 & 20.22 & 7.34 & 27.56 & \cellcolor[HTML]{B8E3B2}72.48 & \cellcolor[HTML]{B8E3B2}36.20 & \cellcolor[HTML]{B7D4EA}0.36 \\
     & Germane & 69 & 91 & 3.61 & 4.67 & 9.00 & 5.28 & 14.28 & \cellcolor[HTML]{63BC6E}32.52 & \cellcolor[HTML]{83CB82}26.42 & \cellcolor[HTML]{9CC9E1}0.40 \\
    \bottomrule
  \end{tabular}%
  }
\end{table}

%% file: tables/cognitive_load_svamp_overview_7.tex
\begin{table}[p]
  \centering
  \caption{Cognitive load summary on SVAMP across five LLMs on Pixel 8 Pro. Within each model block, the energy columns are shaded green (darker = lower / better) and the accuracy column is shaded blue (darker = higher / better).}
  \label{tab:cognitive-load-svamp-overview}
  \tiny
  \setlength{\tabcolsep}{2.5pt}
  \resizebox{\textwidth}{!}{%
  \begin{tabular}{llrrrrrrrrrr}
    \toprule
    Model & Load & Prompt Tokens & Completion Tokens & Prefill Power (W) & Decode Power (W) & Prefill Latency (s) & Decode Latency (s) & Total Latency (s) & Prefill Energy (J) & Decode Energy (J) & Accuracy \\
    \midrule
    \multirow{4}{*}{Gemma-2-2B} & Base & 45 & 88 & 4.20 & 6.40 & 6.21 & 8.31 & 14.52 & \cellcolor[HTML]{63BC6E}25.96 & \cellcolor[HTML]{63BC6E}53.69 & \cellcolor[HTML]{9CC9E1}0.68 \\
     & Intrinsic & 77 & 90 & 4.47 & 6.44 & 7.55 & 8.61 & 16.16 & \cellcolor[HTML]{9FD899}33.62 & \cellcolor[HTML]{83CB82}55.67 & \cellcolor[HTML]{5DA5D1}0.80 \\
     & Extraneous & 128 & 140 & 4.77 & 6.45 & 9.69 & 14.19 & 23.88 & \cellcolor[HTML]{B8E3B2}46.20 & \cellcolor[HTML]{B8E3B2}91.98 & \cellcolor[HTML]{B7D4EA}0.40 \\
     & Germane & 46 & 122 & 4.17 & 6.51 & 6.28 & 11.80 & 18.08 & \cellcolor[HTML]{83CB82}26.12 & \cellcolor[HTML]{9FD899}77.56 & \cellcolor[HTML]{7CB7DA}0.72 \\
    \midrule
    \multirow{4}{*}{LLaMA-3.2-1B} & Base & 61 & 100 & 3.77 & 5.51 & 37.72 & 6.10 & 43.83 & \cellcolor[HTML]{63BC6E}142.21 & \cellcolor[HTML]{83CB82}34.76 & \cellcolor[HTML]{7CB7DA}0.76 \\
     & Intrinsic & 92 & 97 & 3.78 & 5.59 & 56.37 & 5.94 & 62.31 & \cellcolor[HTML]{9FD899}212.80 & \cellcolor[HTML]{63BC6E}34.03 & \cellcolor[HTML]{5DA5D1}0.80 \\
     & Extraneous & 140 & 199 & 3.85 & 5.81 & 82.94 & 12.76 & 95.70 & \cellcolor[HTML]{B8E3B2}319.15 & \cellcolor[HTML]{B8E3B2}78.16 & \cellcolor[HTML]{B7D4EA}0.24 \\
     & Germane & 60 & 133 & 3.76 & 5.81 & 38.37 & 8.17 & 46.54 & \cellcolor[HTML]{83CB82}144.13 & \cellcolor[HTML]{9FD899}48.78 & \cellcolor[HTML]{9CC9E1}0.56 \\
    \midrule
    \multirow{4}{*}{Qwen-2.5-0.5B} & Base & 56 & 118 & 3.71 & 5.11 & 14.48 & 7.01 & 21.49 & \cellcolor[HTML]{83CB82}53.59 & \cellcolor[HTML]{83CB82}39.04 & \cellcolor[HTML]{9CC9E1}0.64 \\
     & Intrinsic & 88 & 115 & 3.64 & 5.41 & 21.57 & 6.78 & 28.35 & \cellcolor[HTML]{9FD899}78.48 & \cellcolor[HTML]{63BC6E}37.88 & \cellcolor[HTML]{5DA5D1}0.76 \\
     & Extraneous & 139 & 318 & 3.66 & 5.76 & 31.98 & 20.94 & 52.92 & \cellcolor[HTML]{B8E3B2}116.91 & \cellcolor[HTML]{B8E3B2}121.37 & \cellcolor[HTML]{B7D4EA}0.29 \\
     & Germane & 56 & 173 & 3.74 & 5.68 & 13.91 & 10.32 & 24.23 & \cellcolor[HTML]{63BC6E}51.91 & \cellcolor[HTML]{9FD899}59.58 & \cellcolor[HTML]{7CB7DA}0.67 \\
    \midrule
    \multirow{4}{*}{Qwen-2.5-1.5B} & Base & 56 & 141 & 3.42 & 5.20 & 44.02 & 13.89 & 57.91 & \cellcolor[HTML]{83CB82}150.46 & \cellcolor[HTML]{83CB82}70.80 & \cellcolor[HTML]{5DA5D1}0.76 \\
     & Intrinsic & 88 & 112 & 3.49 & 5.11 & 64.32 & 11.37 & 75.69 & \cellcolor[HTML]{9FD899}224.40 & \cellcolor[HTML]{63BC6E}56.86 & \cellcolor[HTML]{5DA5D1}0.76 \\
     & Extraneous & 139 & 264 & 3.55 & 4.90 & 99.14 & 29.75 & 128.89 & \cellcolor[HTML]{B8E3B2}352.72 & \cellcolor[HTML]{B8E3B2}142.60 & \cellcolor[HTML]{B7D4EA}0.56 \\
     & Germane & 56 & 182 & 3.40 & 5.23 & 42.98 & 18.34 & 61.33 & \cellcolor[HTML]{63BC6E}146.29 & \cellcolor[HTML]{9FD899}93.25 & \cellcolor[HTML]{8CC0DD}0.64 \\
    \midrule
    \multirow{4}{*}{SmolLM2-360M} & Base & 67 & 154 & 3.69 & 4.89 & 9.15 & 9.07 & 18.22 & \cellcolor[HTML]{83CB82}33.74 & \cellcolor[HTML]{9FD899}45.73 & \cellcolor[HTML]{7CB7DA}0.28 \\
     & Intrinsic & 99 & 110 & 3.67 & 4.85 & 13.18 & 6.52 & 19.70 & \cellcolor[HTML]{9FD899}48.35 & \cellcolor[HTML]{63BC6E}32.09 & \cellcolor[HTML]{5DA5D1}0.36 \\
     & Extraneous & 150 & 179 & 3.63 & 4.88 & 20.25 & 10.55 & 30.80 & \cellcolor[HTML]{B8E3B2}73.33 & \cellcolor[HTML]{B8E3B2}54.44 & \cellcolor[HTML]{B7D4EA}0.12 \\
     & Germane & 67 & 120 & 3.69 & 4.79 & 8.83 & 7.11 & 15.94 & \cellcolor[HTML]{63BC6E}32.60 & \cellcolor[HTML]{83CB82}35.31 & \cellcolor[HTML]{9CC9E1}0.20 \\
    \bottomrule
  \end{tabular}%
  }
\end{table}